\documentclass{article}

\usepackage{arxiv}

\usepackage[utf8]{inputenc} 
\usepackage[T1]{fontenc}    
\usepackage{hyperref}       
\usepackage{url}            
\usepackage{booktabs}       
\usepackage{amsfonts}       
\usepackage{nicefrac}       
\usepackage{microtype}      
\usepackage{lipsum}
\usepackage{graphicx}
\usepackage[table,xcdraw]{xcolor}
\usepackage{amsmath}
\usepackage{bbding}
\usepackage{svg}
\usepackage{subcaption}
\usepackage{algorithm}
\usepackage{algorithmic}
\usepackage{graphicx} 
\usepackage{makecell}
\usepackage{rotating}
\usepackage{adjustbox}

\usepackage{hyperref}
\usepackage{cleveref}

\title{Video-Text Dataset Construction from Multi-AI Feedback: Promoting Weak-to-Strong Preference Learning for Video Large Language Models}

\author{
 Hao Yi \\
 Kuaishou Technology, Beijing, China \\
  Remin University of China,\\Gaoling School of Artificial Intelligence, Beijing \\
  \texttt{yihao@ruc.edu.cn} \\
   \And
 Qingyang Li \thanks{Corresponding author.}\\
  Kuaishou Technology, Beijing, China \\
  \texttt{liqingyang@kuaishou.com} \\
  \And
 Yulan Hu \\
  Kuaishou Technology, Beijing, China \\
  Remin University of China, \\Gaoling School of Artificial Intelligence, Beijing \\
  \texttt{huyulan@kuaishou.com} \\
  \And
Fuzheng Zhang \\
  Kuaishou Technology, Beijing, China \\
  \texttt{zhangfuzheng@kuaishou.com} \\
  \And
Di Zhang \\
  Kuaishou Technology, Beijing, China \\
  \texttt{zhangdi08@kuaishou.com} \\
  \And
Yong Liu \\
  Remin University of China,\\ Gaoling School of Artificial Intelligence, Beijing \\
  \texttt{liuyong@ruc.edu.cn} \\
}

\hypersetup{
pdftitle={A template for the arxiv style},
pdfsubject={q-bio.NC, q-bio.QM},
pdfauthor={David S.~Hippocampus, Elias D.~Striatum},
pdfkeywords={First keyword, Second keyword, More},
}

    \makeatletter
\def\@fnsymbol#1{\ensuremath{\ifcase#1\or \dagger\or \ddagger\or
   \mathsection\or \mathparagraph\or \|\or **\or \dagger\dagger
   \or \ddagger\ddagger \else\@ctrerr\fi}}
    \makeatother
    
\begin{document}
\maketitle
\begin{abstract}

High-quality video-text preference data is crucial for Multimodal Large Language Models (MLLMs) alignment. However, existing preference data is very scarce. Obtaining VQA preference data for preference training is costly, and manually annotating responses is highly unreliable, which could result in low-quality pairs. Meanwhile, AI-generated responses controlled by temperature adjustment lack diversity. To address these issues, we propose a high-quality VQA preference dataset, called \textit{\textbf{M}ultiple \textbf{M}ultimodal \textbf{A}rtificial \textbf{I}ntelligence \textbf{P}reference Datasets in \textbf{V}QA} (\textbf{MMAIP-V}), which is constructed by sampling from the response distribution set and using an external scoring function for response evaluation. Furthermore, to fully leverage the preference knowledge in MMAIP-V and ensure sufficient optimization, we propose \textit{\textbf{Iter}ative \textbf{W}eak-to-\textbf{S}trong \textbf{R}einforcement \textbf{L}earning from \textbf{AI} \textbf{F}eedback for video MLLMs} (\textbf{Iter-W2S-RLAIF}), a framework that gradually enhances MLLMs' alignment capabilities by iteratively updating the reference model and performing parameter extrapolation. Finally, we propose an unbiased and information-complete evaluation scheme in VQA evaluation. Experiments demonstrate that MMAIP-V is beneficial for MLLMs in preference learning and Iter-W2S-RLAIF fully exploits the alignment information in MMAIP-V. We believe that the proposed automatic VQA preference data generation pipeline based on AI feedback can greatly promote future work in the MLLMs alignment. \textbf{Code and dataset are available} \href{https://anonymous.4open.science/r/MMAIP-V_Iter-W2S-RLAIF-702F}{MMAIP-V\_Iter-W2S-RLAIF-702F}.


\end{abstract}

\section{Introduction}~\label{sec:intro}


Recently, Multimodal Large Language Models (MLLMs) have attracted extensive research interest ~\cite{flamingo,qwen2-vl,qwenvl,llava-next-interleave,minicpm,video-chatgpt,video-llava,languagebind}. To improve the instruction-following capability of MLLMs, supervised fine-tuning (SFT) of MLLMs is necessary ~\cite{qwenvl,hound-dpo,palm-e,sft-MLLM}. Howerver, MLLMs after SFT do not align well with human or AI preferences and can generate hallucinations~\cite{hallucination1,hallucination2,gpt4o}. To mitigate the hallucination problem and improve the Video Question \& Answer (VQA) capability in MLLMs, further human or AI feedback alignment is required. In MLLMs alignment, preference learning paradigms similar to that used in natural language processing (NLP) ~\cite{self-reward,dpo,rso,simpo} are adopted ~\cite{rlhf-v,hound-dpo,silkie,vlm-rlaif,hallucination1}, which combines Bradley-Terry model~\cite{bt-model} with the reward implicitly incorporated into the policy model and directly optimizes preference object.

\begin{table}[ht]
\centering
\resizebox{0.5\textwidth}{!}{
\begin{tabular}{c|cccc}
\toprule
 &
  \textbf{\adjustbox{angle=60,valign=c}{ Modal}} &
  \textbf{\adjustbox{angle=60,valign=c}{ Data Format}} &
  \textbf{\adjustbox{angle=60,valign=c}{ Multi-AI Feedback?}} &
  \textbf{\adjustbox{angle=60,valign=c}{ Vision Evaluation?}} \\ \toprule
\textbf{\adjustbox{angle=30,valign=c}{ VLFeedback~\cite{silkie}}} & Image-Text & $(v,x,y_w,y_l)$ & \CheckmarkBold & \CheckmarkBold \\
\textbf{\adjustbox{angle=30,valign=c}{ H-DPO-17k~\cite{hound-dpo}}}  & Video-Text & $(v,x,y_w,y_l)$ & \XSolidBrush   & \XSolidBrush   \\
\textbf{\adjustbox{angle=30,valign=c}{VIDAL~\cite{languagebind}}}    & Video-Text & $(v,x,y)$       & \XSolidBrush   & \XSolidBrush   \\
\textbf{\adjustbox{angle=30,valign=c}{VLM-RLAIF~\cite{vlm-rlaif}}}  & Video-Text & $(v,x,y_w,y_l)$ & \XSolidBrush & \XSolidBrush \\
\textbf{\adjustbox{angle=30,valign=c}{MMAIP-V(ours)}}   & Video-Text & $(v,x,y_w,y_l)$ & \CheckmarkBold & \CheckmarkBold \\ \bottomrule
\end{tabular}
}
\caption{Present vision-text datasets. $v,x,y$ represent the vision input, question and answer, respectively. $y_w, y_l$ represent the chosen answer and the rejected answer.}
\label{tab:datasets}
\end{table}


Although there have been preliminary studies in this field, preference datasets available for MLLMs preference learning are still very limited. The existing relevant open-source datasets are ~\cite{hound-dpo,silkie,languagebind,vlm-rlaif}, shown in \cref{tab:datasets}, and they also suffer from the following issues:  1) Difficulty in collecting preference datasets. Due to the lack of a reasonable and unbiased evaluation model in vision, it is expensive and time-consuming to annotate each video question-answering pair. 2) Due to the varying skill levels and abilities of annotators, the annotation results exhibit a large variance. An automatic generation and evaluation framework \cite{vlm-rlaif, hound-dpo} is required to obtain a large volume of high-quality preference responses. 3) For AI-generated responses, existing methods primarily rely on the sampling randomness introduced by temperature \cite{hound-dpo}. However, this generation method results in similar preference signals, which is not conducive to preference learning and cannot guarantee the high-quality of positive responses and the diversity of negative responses \cite{silkie,vlm-rlaif}, which are crucial properties in VQA preference learning. High-quality preference data has become a significant bottleneck in MLLMs alignment research.

To address above issues from data perspective, we propose an automatic VQA preference data generation pipeline based on multiple AI feedback to generate efficient VQA preference data for MLLMs DPO training, called \textit{\textbf{M}ultiple \textbf{M}ultimodal \textbf{A}rtificial \textbf{I}ntelligence \textbf{P}reference Datasets in \textbf{V}QA} (MMAIP-V), with a total of 24k entries. Specifically, we sample from the response distribution set of well-aligned MLLMs and utilize fine-grained external scoring functions to evaluate response quality. Based on the relative scores from these evaluations, we can construct preference responses pairs.  This data construction pipeline distills effective alignment information from multiple distribution set into the responses, enhancing the diversity of alignment signals in MMAIP-V. Additionally, it uses an external scoring function to reduce biases and noise generated during the distillation process and improve the quality of the alignment signals in MMAIP-V.

From a training perspective, the effective preference signals in the datasets may not be fully exploited due to the limited ability of the reference model. We employ iterative direct preference optimization (DPO) \cite{self-reward} by gradually updating  the reference model, along with a training-free parameters extrapolation \cite{expo} method. These methods fully exploit the potential of preference data, enabling MLLMs to achieve stronger alignment capabilities. We call this preference learning framework as \textit{\textbf{Iter}ative \textbf{W}eak-to-\textbf{S}trong \textbf{R}einforcement \textbf{L}earning from \textbf{AI} \textbf{F}eedback for video MLLMs} (Iter-W2S-RLAIF). Furthermore, from an evaluation perspective,  we argue that previous evaluations are biased and lack vision information by matching only the ground truth and responses while ignoring vision information. To evaluate MLLMs more comprehensively and with less bias, we incorporate video information to conduct fine-grained, multi-perspective evaluations.

We evaluate our model on three in-domain and four out-domain test datasets. The experimental results demonstrate that the high quality of the positive responses and the diversity of the negative responses in MMAIP-V are beneficial for preference learning in MLLMs. Furthermore, the iterative preference learning paradigm combined with the training-free parameters extrapolation method, effectively and fully leverages the AI preference feedback contained in MMAIP-V. Our contributions are as follows:
\begin{figure*}[htbp]
	\centering
	\begin{minipage}[c]{0.49\textwidth}
		\centering
		\includegraphics[width=1\textwidth]{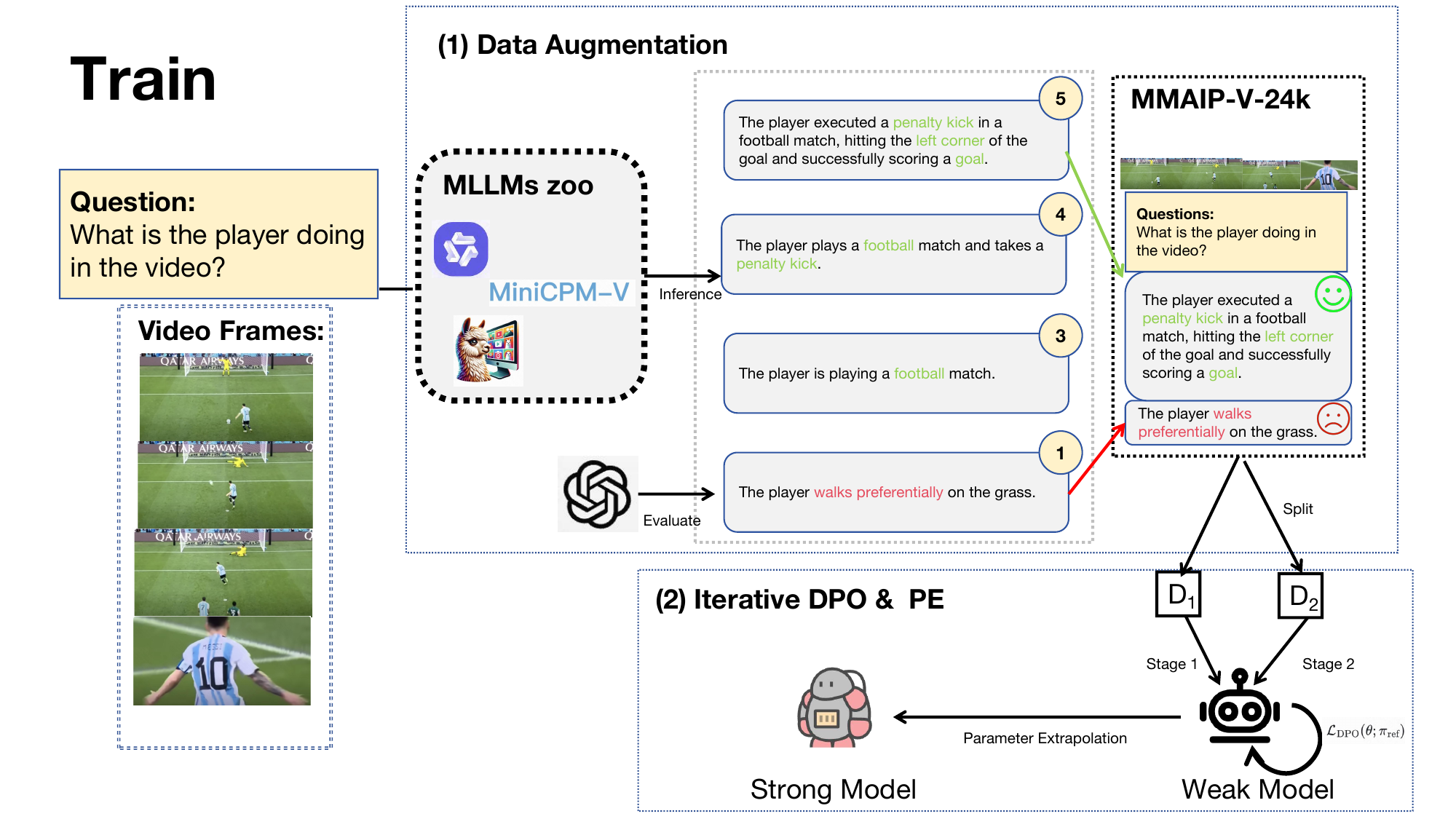}
		\subcaption{VQA train framework.}
		\label{fig:train-framework}
	\end{minipage} 
	\begin{minipage}[c]{0.49\textwidth}
		\centering
		\includegraphics[width=\textwidth]{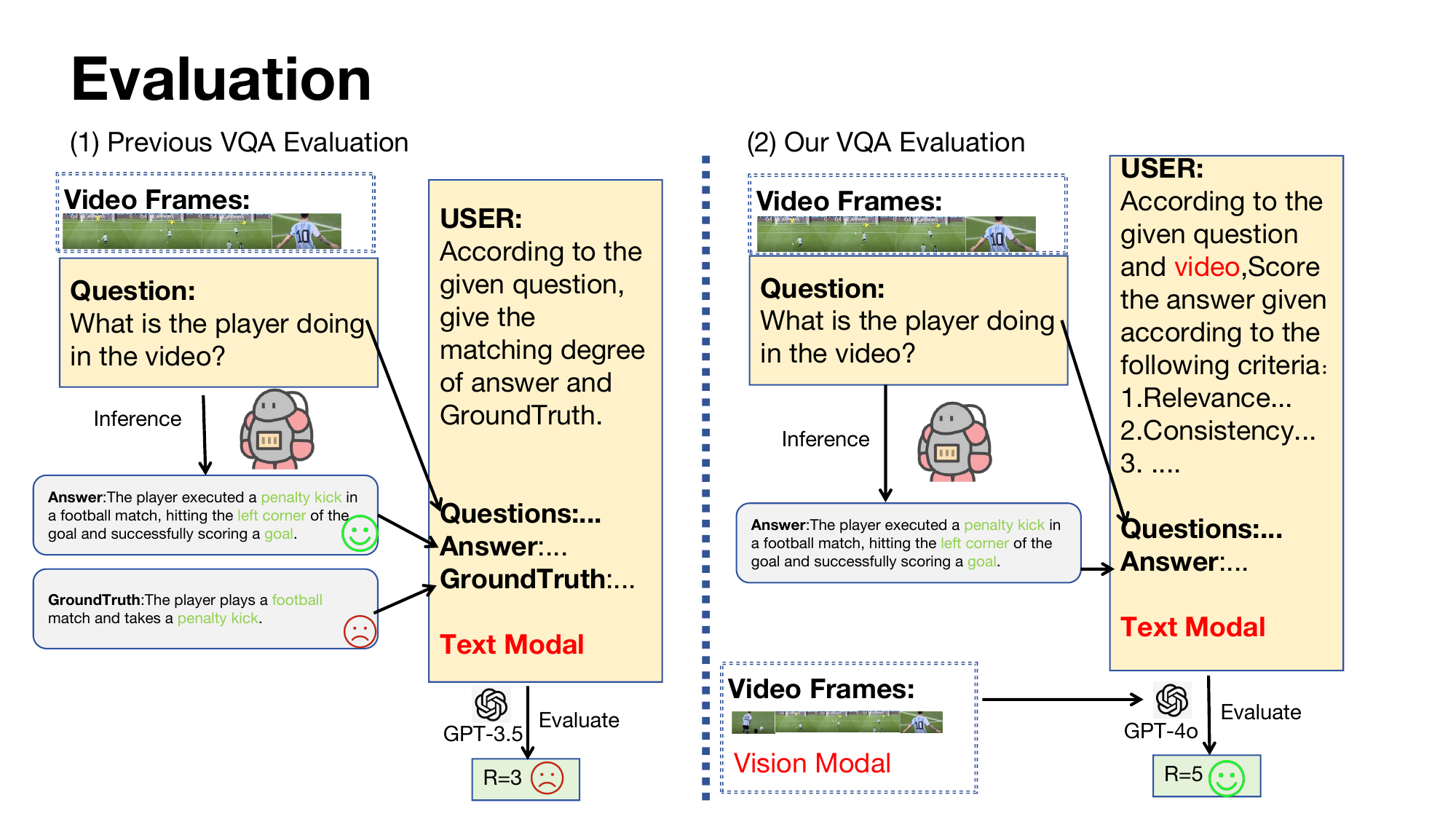}
		\subcaption{VQA evaluation framework. }
		\label{fig:eval-framework}
	\end{minipage} 
	\caption{Train and Evaluation framework. \textbf{(a)} we first constructs VQA preference datasets by sampling from a diverse and high-quality distribution of MLLMs response (MLLMs zoo) and evaluating the sampled responses to select chosen response and rejected response. Subsequently, the dataset is divided, and iterative DPO and parameters extrapolation are performed. The model's alignment capability gradually improves. \textbf{(b)} Difference between previous VQA evaluation method and our proposed VQA evaluation method. Previous method solely relies on MLLMs responses matching groundtruth and is lack of vision information. Our proposed method incorporates video information into the evaluation process and evaluate MLLMs responses from multiple perspectives. }

	\label{fig:framework}
\end{figure*}

\begin{itemize}
    \item We construct a high-quality and diverse VQA preference dataset by sampling from the response distribution set, called MMAIP-V. Experiments demonstrate that MMAIP-V is beneficial for preference alignment in MLLMs.
    
     \item  We propose Iter-W2S-RLAIF, an iterative weak-to-strong reinforcement learning paradigm based on AI feedback. By fully leveraging AI preference signals, we improve VQA generation capability of MLLMs.
    \item We propose a vision-based and unbiased evaluation scheme, which eliminates the biases and vision information loss inherent in previous evaluations.
\end{itemize}

\section{Related Work}
\label{sec:rel-work}

\textbf{Vision-text Datasets}
There are serval works to construct high quality video-text datasets \cite{languagebind,actnet,webvid}. For example, VIDAL \cite{languagebind}  contains 10 million data with video and its corresponding language. All videos are from short video platforms with complete semantics rather than truncated segments from long videos. However, there is lack of high-quality and  public VQA  datasets for preference learning. \cite{hound-dpo} utilizes captions of video and GPT-3.5 to  construct total 17k vedio-text preference pair,  but the quality of it is low and is not suitable for better preference learning. Silkie \cite{silkie} utilizes Large Vision Language Model pool to construct image-text  preference pair, called VLFeedback. But VLFeedback is not in video-text fields. The lack of open and high-quality video-text preference datasets for effective DPO training motivates this work.

\textbf{Multimodal Large Language Models.} Recently, Multimodal Large Language Models (MLLMs) have attracted extensive research interest \cite{llava-next-interleave,video-chatgpt,llama-vid,llavanextvideo,video-llava,videollama2}. MLLMs utilize  Large Language Models (LLMs) to strength the multimodal comprehension. Specifically in video modal understanding,  MLLMs  mostly utilize a pretrained vision encoder like CLIP \cite{clip} or LanguageBind \cite{languagebind} to encode the vision information. Video MLLMs is pretrained  in a large video-text pair \cite{hound-dpo,video-llava,languagebind}. Then MLLMs undergo the Supervised Fine-Tuning (SFT) to promote the MLLMs capability of instruction following.

\textbf{Preference Alignment in MLLMs.}
MLLMs after SFT do not align well with human preferences and can generate hallucinations. To address this issue, previous studies mostly incorporate Reinforcement Learning from Human Feedback (RLHF) and DPO, aiming to inject human or AI feedback information into MLLMs ~\cite{rlhf-v,hound-dpo,silkie,vlm-rlaif,hallucination1}.  RLHF-V \cite{rlhf-v} enhances MLLM trustworthiness via behavior alignment from fine-grained correctional human feedback. VLM-RLAIF \cite{vlm-rlaif} employs multimodal AI system to enhance itself. LLaVA-Hound-DPO \cite{hound-dpo} utilizes detailed video captions as a proxy of video content, enabling language models to incorporate this information as supporting evidence for scoring video Question Answering (QA) predictions. Silkie \cite{silkie} distillates image preference from vision Large Language pool and adopt GPT-4V to assess the generated outputs regarding helpfulness, visual faithfulness, and ethical considerations.

\section{MMAIP-V}
\label{sec:MMAIP-V}

 In this section, we introduce the construction pipeline of MMAIP-V. 
 
\subsection{Multiple Multimodal Artificial Intelligence Preference Data Augmentation in VQA }
\label{sec:method-model-zoo}
To mitigate the challenges of collecting VQA preference datasets and the high costs associated with annotation, we employ multiple visual language models feedback and utilize scores from a visual reward model to automatically construct VQA preference datasets. Specifically, we have a video MLLMs zoo $\mathbf{M}=\{\mathcal{M}_i\}_{i=1}^K$, where $\mathcal{M}_i:\mathcal{V}\times \mathcal{X}\rightarrow\mathcal{Y}$, and $\mathcal{V}$, $\mathcal{X}$,  $\mathcal{Y}$ is the distribution of videos, prompts and responses, respectively. We construct the VQA preference dataset as follow:
\begin{align}
    \mathcal{D}^{VT}_{DPO}&=\{(v_i,x_i,y_w,y_l)\}_{i=1}^N, \nonumber \\
    y_w&=\arg\max_{y\sim\mathbf{M}}R(y;x,v)\nonumber, \\
    y_w&=\arg\min_{y\sim\mathbf{M}}R(y;x,v),
    \label{equ:vt-datasets}
\end{align}
where $N$ is total amount of dataset, $v_i$, $x_i$ are sampled from $\mathcal{V}$, $\mathcal{X}$. $R: \mathcal{V}\times\mathcal{X}\times\mathcal{Y} \rightarrow \mathbb{R}$ is a visual reward model that evaluates visual information and its related question-answer pairs from multiple perspectives  at a fine-grained level. A higher score indicates that the model's response is more relevant to the visual information and the question, which means fewer hallucinations are generated. Vision reward model gives the AI feedback for preference learning. In practice, we drop the video-question pair which has the same score between all candidate responses to exclude the useless question for preference learning. Due to the lack of trained reliable reward model, we utilize GPT-4o to score responses in a text-to-text manner.  A toy example is shown in the Data Augmentation part of \cref{fig:train-framework}.

\subsection{Detailed Construction Pipeline } 
\textbf{Video-Question Pair Construction.}  For the training data, we integrated a total of 900k video question-answer pairs from the WebVid (400k)~\cite{webvid}, VIDAL (450k)~\cite{languagebind} and ActivityNet (50k)~\cite{actnet} datasets, and uniformly sampled 30k as our training video-question source. WebVid and VIDAL are in the general domain sourced from YouTube, and ActivityNet videos focus on human activities. WebVid, Vidal, and ActivityNet account for 33\%, 32\%, and 34\%, respectively.

\textbf{Sampling Responses from MLLMs Zoo.} To ensure the high quality of positive preference data and the diversity of responses from the MLLMs zoo, we selected models with varying model sizes, architectures, and capabilities: \texttt{Qwen2-VL-72B-Instruct}~\cite{qwen2-vl}, \texttt{MiniCPM-V-2.6}~\cite{minicpm}, \texttt{Qwen2-VL-7B-Instruct}~\cite{qwen2-vl}, \texttt{Qwen2-VL-2B-Instruct}~\cite{qwen2-vl}. For the 30k training video-question pairs constructed as mentioned above, we utilize models in the MLLMs zoo to inference, setting the inference temperature to $1.0$. 

\textbf{Automatic Scoring.} After obtaining the responses from the various vedio MLLMs,  we use \texttt{GPT-4o}~\cite{gpt4o} to comprehensively score the responses based on relevance, consistency, accuracy, specificity, comprehensiveness, and novel insight, with the evaluation scores ranging from $\{1,2,3,4,5\}$. The detailed design of the scoring prompts refers to  \cref{fig:infer_reward}.

\textbf{Responses Pair Construction.} For each video-question pair, after obtaining multiple responses and filtering out questions where all responses received the same score, we select the highest-scoring response as the positive response for that question and the lowest-scoring response as the negative response. In total, we obtain 24k Video Question\&Answering preference responses pairs.

\subsection{ Details on MMAIP-V}

\textbf{GPT-4o score distributions of responses in the MLLMs zoo.} To reflect the quality of responses generated by various MLLMs models, we  plot the distribution of GPT-4o evaluation scores for each model in MMAIP-V, including Qwen2-VL-72B-Instruct, MiniCPM-V-2.6, Qwen2-VL-7B-Instruct, Qwen2-VL-2B-Instruct, referring to \cref{fig:score-distribution}. It can be seen that Qwen2-VL-72B-Instruct has the highest average score on MMAIP-V, which is 4.82, while Qwen2-VL-2B-Instruct has the lowest average score, which is 3.88. The weaker the model alignment capability, the more long-tailed the score distribution. The response score distributions of Qwen2-VL-72B-Instruct and MiniCPM-V-2.6 are more concentrated on high score, ensuring high quality of positive responses, while the score distributions of Qwen2-VL-7B-Instruct and Qwen2-VL-2B-Instruct are more long-tailed, ensuring the diversity of negative responses.

\begin{figure}[ht]
	
    \centering
    \includegraphics[width=0.8\textwidth]{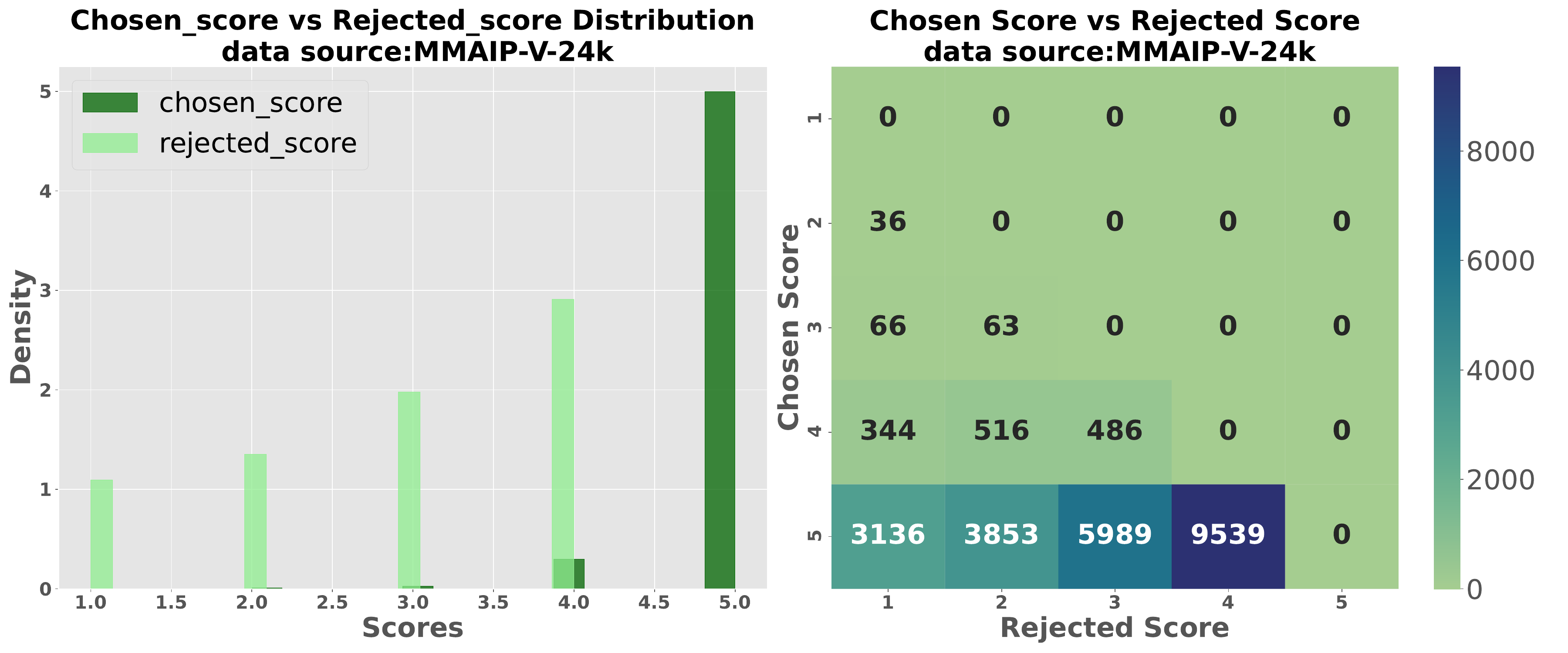}

\caption{ Left: Joint score distribution of positive and negative responses pairs in MMAIP-V. The row represents the rejected responses score and the column represents the chosen responses score. Right: Marginal score distribution of positive and negative responses pairs in MMAIP-V.The dark green bar represents the chosen responses distribution and the light green bar represent the rejected responses distribution.}

\label{fig:score-distribution—pair}
\end{figure}

\textbf{Score distribution of MMAIF-V pairs.} We analyze the joint and marginal distributions of positive and negative response scores on MMAIP-V, as shown in  \cref{fig:score-distribution—pair}. The joint distribution of positive and negative response scores demonstrates that MMAIP-V has a certain number of samples across various score combinations, with approximately 40\% of samples concentrated in a score distribution of (5,4) and roughly 25\% of samples concentrated in a score distribution of (5,3). We believe that hard negative responses are crucial to the effectiveness of MMAIP-V in VQA preference alignment. The marginal distribution of positive and negative response scores reveals that most positive response scores are concentrated at score 5, with a smaller portion at score 4, while the distribution of negative response scores is more uniform. This indicates that better positive responses and more diverse negative responses are beneficial for VQA preference alignment. More details about MMAIP-V are demonstrated in  \cref{sec:MMAIP-V-datails}.

\section{Iter-W2S-RLAIF}
In this section, to fully utilize the AI preference alignment signals in MMAIP-V, we propose Iter-W2S-RLAIF. \cref{sec:method-dpo} introduces preference learning in the context of VQA tasks and iterative DPO.  \cref{sec:expo} introduces parameters extrapolation in MLLMs.

\subsection{Iterative Direct Preference Optimization in VQA}
\label{sec:method-dpo}
\paragraph{Direct Preference Optimization (DPO)~\cite{dpo}.} It has been extensively employed in the LLMs alignment and has recently used in the MLLMs alignment~\cite{hound-dpo,silkie,hall-dpo}. By addressing the reinforcement learning objectives, the reward is implicitly incorporated into the policy model. When combined with the Bradley-Terry model~\cite{bt-model}, DPO facilitates the offline reinforcement paradigm to directly align AI preference. Specifically, after we obtain the VQA preference datasets $\mathcal{D}_\text{DPO}^{VT}$ \ref{equ:vt-datasets}, DPO minimize the preference loss function defined as follow:
\begin{align}
    \mathcal{L}_\text{DPO}(\theta; \pi_\text{ref})=\mathbb{E}_{(v,x,y_w,y_l)\sim\mathcal{D}_\text{DPO}^{VT}}[-\log\sigma(\hat{r}_{\theta}(y_w)-\hat{r}_{\theta}(y_l))],
    \label{equ:dpo-loss}
\end{align}
where $\hat{r}_{\theta}(y) = \beta\log\frac{\pi_{\theta}(y|v,x)}{\pi_{ref}(y|v,x)}$, $\pi_{\theta}$ is the video multimodal policy model, $\theta$ is its learnable parameters, $\pi_{ref}$ is reference model, which is regarded as a regular term. Loss \cref{equ:dpo-loss} makes MLLMs response distribution as closely as possible to $\mathcal{Y}^w$ while maximizing the distance from $\mathcal{Y}^l$.

\paragraph{Iterative DPO.} This method can explore more parameters space by updating the reference model $\pi_{ref}$ iteratively. \cite{understand-ref} has proved that a stronger reference policy can enhance DPO performance when the reference policy is compatible from theoretical and experimental perspective. Specifically, we first split the $\mathcal{D}^{VT}_{DPO}$ evenly to get $T$ sub-datasets $\{\mathcal{D}_i\}_{i=1}^T$, where $T$ is the iteration time. For step $t$, we minimize the loss \Cref{equ:dpo-loss} on $\mathcal{D}_{t}$, which will get the last policy $\pi_{\theta_{t}}$. Meanwhile, we update the reference model $\pi_{ref}=\pi_{\theta_{t}}$.

\subsection{Parameters Extrapolation.} 
\label{sec:expo}
Recent research has concentrated on parameters mixing and parameters extrapolation methods to enhance the response quality of LLMs~\cite{expo-1,expo-2,expo-3,expo-4}. However, these parameters extrapolation methods have yet to be effectively validated for improving the alignment capabilities of MLLMs. Specifically, if we have two aligned MLLMs $\{\pi_{\theta_1},\pi_{\theta_2}\}$ such that:
\begin{align}
    \mathbb{E}_{y\sim\pi_{\theta_2}}[R(v,x,y)]>\mathbb{E}_{y\sim\pi_{\theta_1}}[R(v,x,y)].
\end{align}
For example, we do one-step DPO based on $\pi_{\theta_0}$ and obtain  $\pi_{\theta_1}$. Then we extend the  ``aligned vector'' $\Delta \theta$, we can get stronger MLLMs $\pi_{\theta^*}$ in alignment capability:
\begin{align}
    \theta^*=\theta_2+\alpha(\theta_2-\theta_1)=\theta_2+\alpha\Delta\theta,
\end{align}
where $\alpha>0$ is the extend strength of parameters extrapolation. This method can promote the capability of MLLMs in a training-free way, which do not utilize any post-training or fine-tune operation.

The integral and detailed pipeline of  MMAIP-V Construction  and  Iter-W2S-RLAIF is shown in  \cref{alg1} and  \cref{fig:framework}.

\section{The Unbiased VQA Evaluation Scheme}
\label{sec:method-eval}

\begin{figure*}[htbp]
	\centering
	\begin{minipage}[c]{0.95\textwidth}
		\centering
    \includegraphics[width=0.75\linewidth]{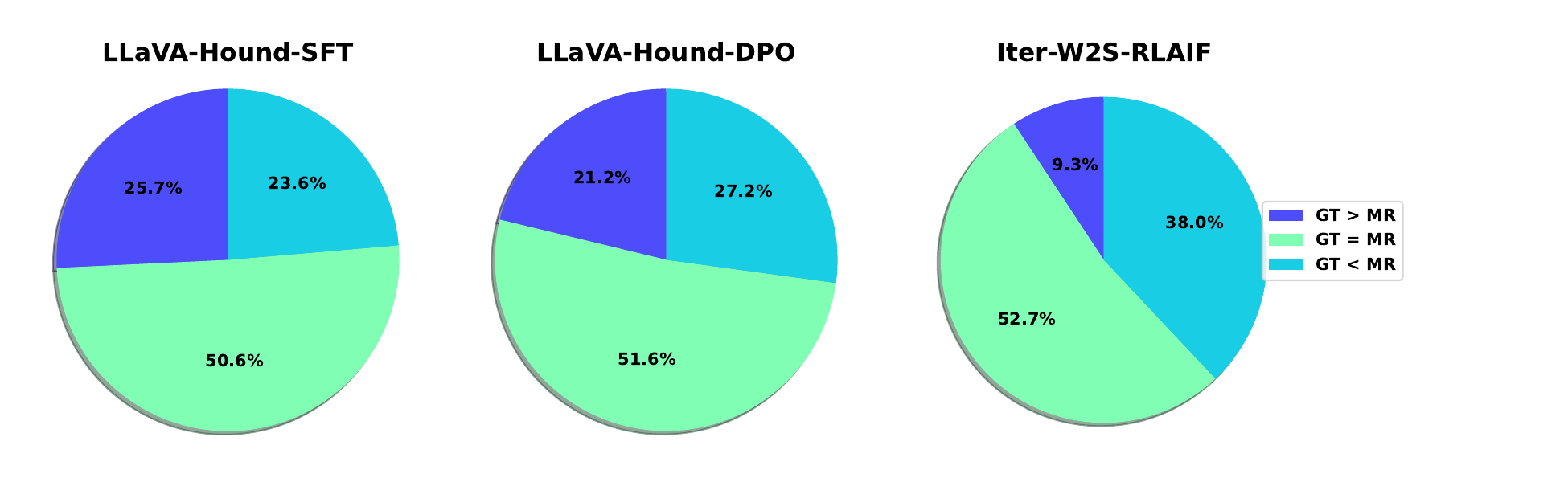}
    \subcaption{Comparison of groundtruth and model response scores.``GT$>$MR'' represents groundtruth being better than the MLLM response;  ``GT$=$MR'' represents groundtruth and the MLLMs response performing.  ``GT$<$MR'' represents groundtruth being worse than the MLLMs response.}
    \label{fig:eval-biased}
	\end{minipage} \\
	\begin{minipage}[c]{0.49\textwidth}
		\centering
		\includegraphics[width=\textwidth]{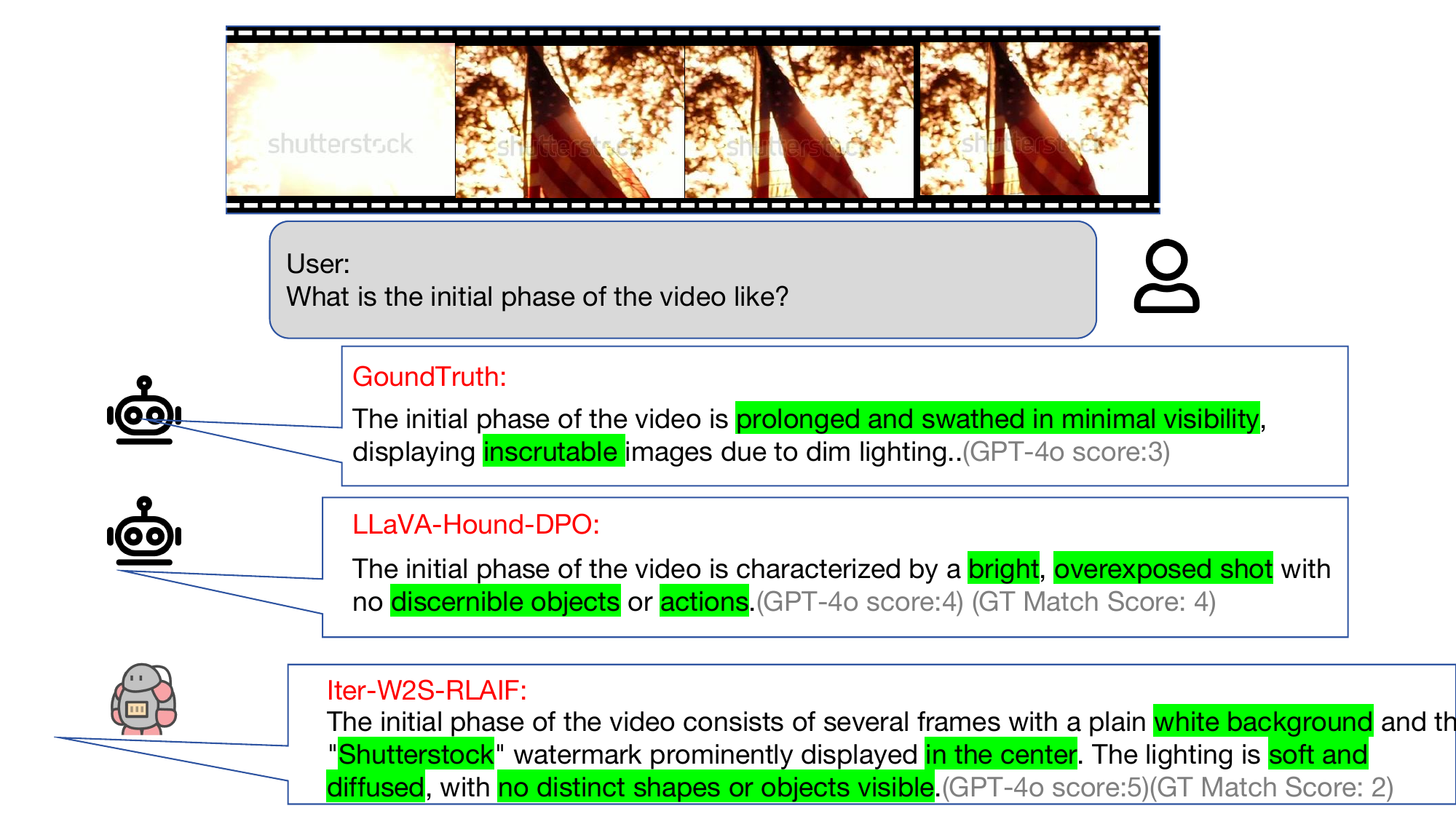}
		\subcaption{Incomplete description.}
		\label{fig:eval-case-1}
	\end{minipage} 
    \begin{minipage}[c]{0.49\textwidth}
		\centering
		\includegraphics[width=\textwidth]{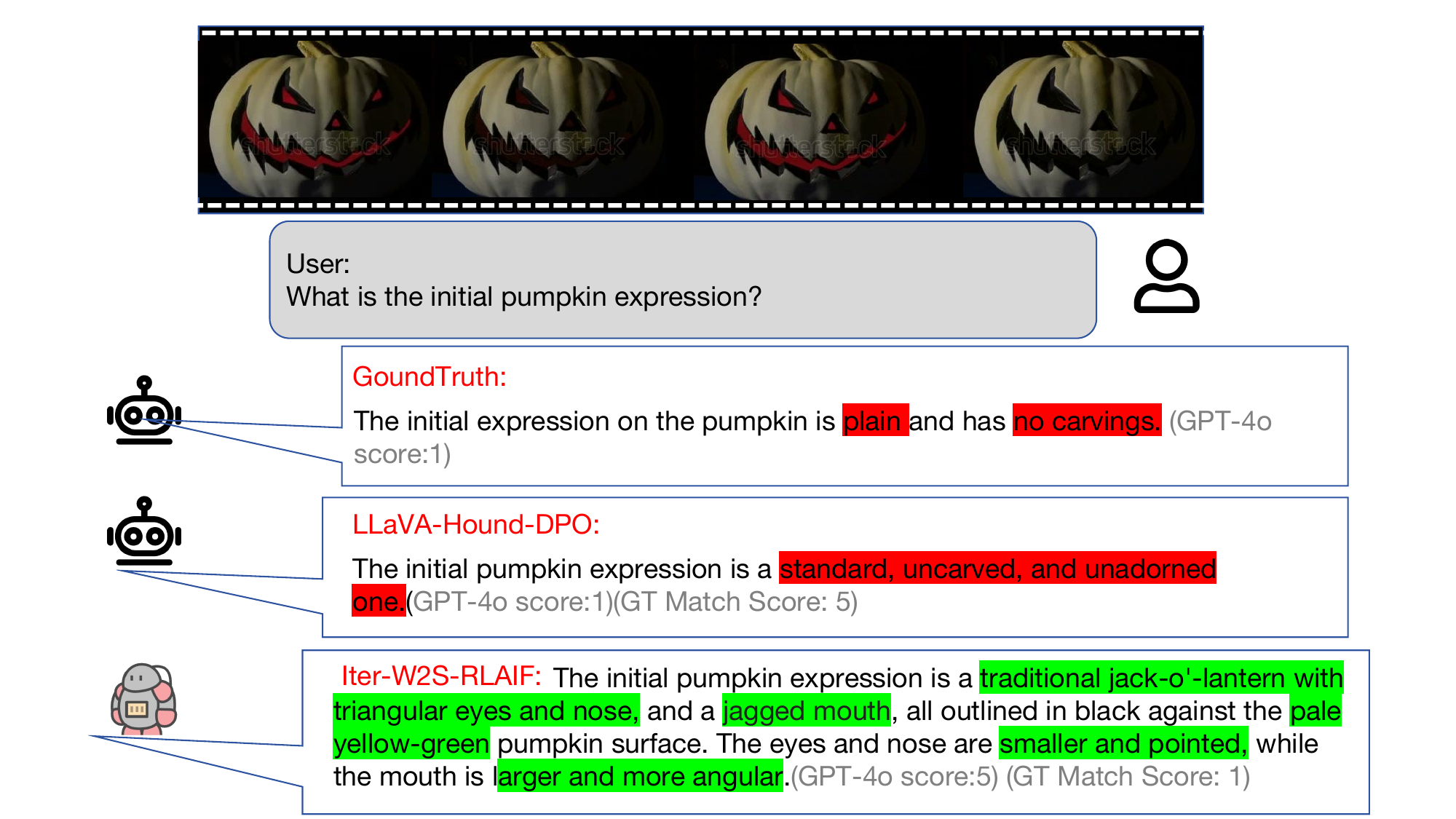}
		\subcaption{Incorrect description.}
		\label{fig:eval-case-2}
	\end{minipage} 
	\caption{Illustration of previous evaluation scheme  drawbacks.}
    \label{fig:evaluation-bias}
\end{figure*}

Previous studies~\cite{video-chatgpt,llavanextvideo,llama-vid,videollama2}  evaluate video MLLMs performance in VQA by comparing how closely the video MLLMs responses match the groundtruth. Detailed evaluation prompts refer to  \cref{fig:old-eval-prompt}. However, this evaluation method has two main drawbacks:\textbf{ 1)} \textit{The evaluation phase lacks the supervision of visual modality information.} \textbf{2)} \textit{Groundtruth might not represent the optimal VQA result.} The above drawbacks lead to biased evaluation in the video question-answering alignment. It can be imagined that if the quality of the inference answers surpasses that of the groundtruth, the evaluation results instead show negative signals. 

There are some experimental phenomenons can illustrate the hypothesis. In \cref{fig:eval-biased}, to illustrate that the evaluation model relying solely on MLLM responses to match groundtruth is biased, we infer responses from three aligned MLLMs (LLaVA-Hound-SFT, LLaVA-Hound-DPO, Iter-W2S-RLAIF (\textbf{ours}), with increasing alignment capabilities on the WebVid test datasets as shown in the  \cref{sec:exp-main}. It can be observed that as the alignment capability of the MLLMs improves, the groundtruth is more unreliable. This validates our hypothesis that the stronger the MLLMs capability, the greater the bias introduced by previous evaluation method, rendering it unable to accurately reflect the alignment performance of the MLLMs. Meanwhile, there are two cases come from the WebVid test datasets. In \cref{fig:eval-case-1}, the groundtruth  is incomplete, and in \cref{fig:eval-case-2}, the groundtruth  contains factual errors. The answers from Iter-W2S-RLAIF are more comprehensive and factually accurate. However, comparing against the groundtruth can lead to biased evaluations. Therefore, we introduce the video modality and perform a fine-grained evaluation of the inference answer quality by utilizing \texttt{GPT-4o}~\cite{gpt4o}, thereby eliminating the bias arising from ``the match degree between MLLM responses and groundtruth'', shown in \cref{fig:eval-framework}. Detailed  evaluation prompts we proposed refer to  \cref{fig:new-eval-prompt}.

\section{Experiment}
\label{sec:exp}
\subsection{ Experimental Setup}
\label{sec:exp-setup}

\textbf{Training Datasets.} To verify the effectiveness of the MMAIP-V construction pipeline in  \cref{sec:MMAIP-V} for VQA preference learning, we use all the preference data from MMAIP-V as the training dataset for VQA preference learning without any additional processing.

\textbf{Model \& Training  Details.} The base model used in this experiment is \texttt{LLaVA-Hound-SFT} \cite{hound-dpo}. For all videos, we uniformly extract 8 frames as input for the  visual encoder. For the iterative DPO training, we set 2 rounds for iterations and evenly split the constructed 24k preference data into two parts, denoted as $\mathcal{D}_1$ and $\mathcal{D}_2$. For parameters extrapolation, we perform parameters extrapolation at the end of both the first and second stages. All experiments are conducted on 8 Nvidia H800 80GB GPUs.

\textbf{Evaluation Detail.} For the evaluation of the model alignment capability and video question-answering ability, we conduct in-domain evaluations (WebVid(5955) \cite{webvid}, VIDAL(5991) \cite{languagebind}, ActivityNet(5898) \cite{actnet}) and out-of-domain evaluations MSRVTT(6090) \cite{msrvtt}, MSVD(4959) \cite{msvd}, TGIF(6000) \cite{tgif} and SSV2 \cite{ssv2}(5999). We evaluate all the MLLMs as discussed in in  \cref{sec:method-eval}. We record the model average score (\textbf{Score}) and the proportion of test datasets with scores greater than or equal to 3 (\textbf{Ratio}). More experimental details refer to \cref{sec:exp-detail}

 \begin{table*}[tph]
\centering
\resizebox{0.95\textwidth}{!}{
\begin{tabular}{@{}cccccccc@{}}
\toprule

\textbf{Method} &
  \textbf{Model} &
  \multicolumn{6}{c}{\textbf{In-Domain}} \\ \midrule

 &
   &
  \multicolumn{2}{c}{\textbf{WebVid}} &
  \multicolumn{2}{c}{\textbf{VIDAL}} &
  \multicolumn{2}{c}{\textbf{ActivityNet}} \\

 &
   &
  Score &
  Ratio &
  Score &
  Ratio &
  Score &
  Ratio \\ \hline

\multicolumn{1}{c|}{} &
  VideoChatGPT\cite{video-chatgpt} &
  3.93 &
  79.26 &
  3.52 &
  70.66 &
  3.75 &
  76.60 \\

\multicolumn{1}{c|}{} &
  LLaMA-VID\cite{llama-vid} &
  3.89 &
  79.81 &
  3.57 &
  73.19 &
  3.88 &
  80.52 \\

\multicolumn{1}{c|}{Pretrained} &
  Video-LLaVA\cite{video-llava} &
  4.21 &
  86.88 &
  3.75 &
  77.00 &
  3.73 &
  75.11 \\

\multicolumn{1}{c|}{} &
  LLaVA-Next-Video-Base\cite{llavanextvideo} &
  4.28 &
  89.00 &
  3.88 &
  80.84 &
  4.13 &
  86.21 \\

\multicolumn{1}{c|}{} &
  Video-LLaMA2-Base\cite{videollama2} &
  4.34 &
  90.87 &
  3.86 &
  81.59 &
  3.98 &
  84.24 \\ \midrule

\multicolumn{1}{c|}{SFT} &
  LLaVA-Hound-SFT\cite{hound-dpo} &
  {\color[HTML]{1F2329} 4.50} &
  {\color[HTML]{1F2329} 93.52} &
  4.27 &
  90.21 &
  4.38 &
  92.71 \\

\multicolumn{1}{c|}{} &
  Video-LLaMA2-Chat\cite{videollama2} &
  3.82 &
  77.14 &
  3.62 &
  74.23 &
  4.13 &
  86.49 \\ \midrule

\multicolumn{1}{c|}{} &
  VLM-RLAIF\cite{vlm-rlaif} &
  4.14 &
  88.58 &
  3.8 &
  80.14 &
  4.04 &
  85.68 \\

\multicolumn{1}{c|}{} &
  LLaVA-Next-Video-DPO\cite{llavanextvideo} &
  {\color[HTML]{1F2329} 4.47} &
  {\color[HTML]{1F2329} 93.17} &
  4.10 &
  86.77 &
  4.30 &
  90.50 \\

\multicolumn{1}{c|}{RLHF/RLAIF} &
  LLaVA-Hound-DPO\cite{hound-dpo} &
  {\color[HTML]{1F2329} 4.55} &
  {\color[HTML]{1F2329} 95.12} &
  4.34 &
  92.42 &
  4.41 &
  94.20 \\

\multicolumn{1}{c|}{} &
  LLaVA-Hound-DPO$\dag$ &
  4.56 &
  94.92 &
  4.36 &
  92.81 &
  4.44 &
  94.62 \\ \cmidrule(l){2-8} 

\multicolumn{1}{c|}{} &
  Iter-W2S-RLAIF(\textbf{Ours}) &
  {\color[HTML]{1F2329} \textbf{4.78}} &
\textbf{ 97.67}&
  {\color[HTML]{2C2F33} \textbf{4.52}} &
 \textbf{  93.66} &
  {\color[HTML]{2C2F33} \textbf{4.64}} &
  {\color[HTML]{2C2F33} \textbf{96.13}} \\
\multicolumn{1}{c|}{} &
  $\Delta_{Base}$ &
  {\color[HTML]{FE0000} +0.28} &
  {\color[HTML]{FE0000} +4.15} &
  {\color[HTML]{FE0000} +0.25} &
  {\color[HTML]{FE0000} +3.45} &
  {\color[HTML]{FE0000} +0.26} &
  {\color[HTML]{FE0000} +3.42} \\ \bottomrule
\end{tabular}

}
\caption{Main results on VQA in-domain generation.We report the Score and Ratio on the three in-domain test datasets. LLaVA-Hound-DPO\dag \quad is the model we reproduced according to the original paper and code~\cite{hound-dpo}. 
$\Delta_{Base}$ represents the improvement of our model Iter-W2S-RLAIF relative to the base model LLaVA-Hound-SFT. All the model parameters size we evaluate is 7B. More baseline models details is demonstrated in  \cref{sec:baselines-detail}. }
\label{tab:main-result-indomain}
\end{table*}

\begin{table*}[htp]
\centering
\resizebox{0.95\textwidth}{!}{
\begin{tabular}{@{}
c 
c 
c 
c 
c 
c 
c 
c 
c 
c @{}}
\toprule
\textbf{Method} &
  \textbf{Model} &
  \multicolumn{8}{c}{\cellcolor[HTML]{FFFFFF}\textbf{Out-Domain}} \\ \midrule
 &
   &
  \multicolumn{2}{c}{\cellcolor[HTML]{FFFFFF}{\color[HTML]{1F2329} \textbf{SSV2}}} &
  \multicolumn{2}{c}{\cellcolor[HTML]{FFFFFF}{\color[HTML]{1F2329} \textbf{MSRVTT}}} &
  \multicolumn{2}{c}{\cellcolor[HTML]{FFFFFF}{\color[HTML]{1F2329} 
 \textbf{MSVD}}} &
  \multicolumn{2}{c}{\cellcolor[HTML]{FFFFFF}{\color[HTML]{1F2329} \textbf{TGIF}}} \\
 &
   &
  Score &
  Ratio &
  Score &
  Ratio &
  Score &
  Ratio &
  Score &
  Ratio \\ \hline
\multicolumn{1}{c|}{\cellcolor[HTML]{FFFFFF}} &
  VideoChatGPT\cite{video-chatgpt} &
  3.50 &
  70.10 &
  3.46 &
  68.16 &
  3.74 &
  75.03 &
  3.91 &
  79.18 \\
\multicolumn{1}{c|}{\cellcolor[HTML]{FFFFFF}} &
  LLaMA-VID\cite{llama-vid} &
  3.66 &
  76.58 &
  3.48 &
  69.38 &
  3.80 &
  77.71 &
  3.81 &
  78.02 \\
\multicolumn{1}{c|}{\cellcolor[HTML]{FFFFFF}Pretrained} &
  Video-LLaVA\cite{video-llava} &
  3.75 &
  78.17 &
  3.62 &
  73.17 &
  3.96 &
  80.86 &
  3.59 &
  71.83 \\
\multicolumn{1}{c|}{\cellcolor[HTML]{FFFFFF}} &
  LLaVA-Next-Video-Base\cite{llavanextvideo} &
  3.84 &
  80.81 &
  3.86 &
  80.08 &
  4.12 &
  85.25 &
  4.20 &
  87.61 \\
\multicolumn{1}{c|}{\cellcolor[HTML]{FFFFFF}} &
  Video-LLaMA2-Base\cite{videollama2} &
  3.58 &
  78.53 &
  3.76 &
  78.59 &
  4.06 &
  85.24 &
  4.07 &
  87.21 \\ \midrule
\multicolumn{1}{c|}{\cellcolor[HTML]{FFFFFF}SFT} &
  LLaVA-Hound-SFT\cite{hound-dpo} &
  {\color[HTML]{1F2329} 4.22} &
  {\color[HTML]{1F2329} 90.72} &
  4.20 &
  87.92 &
  4.41 &
  91.71 &
  4.49 &
  94.60 \\
\multicolumn{1}{c|}{\cellcolor[HTML]{FFFFFF}} &
  Video-LLaMA2-Chat\cite{videollama2} &
  3.85 &
  80.10 &
  3.87 &
  80.89 &
  4.15 &
  85.69 &
  4.20 &
  87.37 \\ \midrule
\multicolumn{1}{c|}{\cellcolor[HTML]{FFFFFF}} &
  VLM-RLAIF\cite{vlm-rlaif} &
  3.73 &
  78.59 &
  3.77 &
  79.38 &
  4.05 &
  84.57 &
  4.16 &
  88.48 \\
\multicolumn{1}{c|}{\cellcolor[HTML]{FFFFFF}} &
  LLaVA-Next-Video-DPO\cite{llavanextvideo} &
  {\color[HTML]{1F2329} 4.07} &
  {\color[HTML]{1F2329} 87.37} &
  4.10 &
  86.59 &
  4.30 &
  90.68 &
  4.36 &
  92.00 \\
\multicolumn{1}{c|}{\cellcolor[HTML]{FFFFFF}RLHF/RLAIF} &
  LLaVA-Hound-DPO\cite{hound-dpo} &
  {\color[HTML]{1F2329} 4.27} &
  {\color[HTML]{1F2329} 94.25} &
  4.24 &
  90.23 &
  4.42 &
  92.87 &
  4.47 &
  93.48 \\
\multicolumn{1}{c|}{\cellcolor[HTML]{FFFFFF}} &
  LLaVA-Hound-DPO$\dag$ &
  4.28 &
  94.24 &
  4.24 &
  89.86 &
  4.44 &
  93.35 &
  4.53 &
  95.60 \\ \cmidrule(l){2-10} 
\multicolumn{1}{c|}{\cellcolor[HTML]{FFFFFF}} &
  Iter-W2S-RLAIF(\textbf{Ours}) &
  {\color[HTML]{1F2329} \textbf{4.51}} &
  {\color[HTML]{1F2329} \textbf{95.48}} &
  {\color[HTML]{2C2F33} \textbf{4.52}} &
  \textbf{93.22} &
  {\color[HTML]{2C2F33} \textbf{4.61}} &
  {\color[HTML]{1F2329} \textbf{94.66}} &
  {\color[HTML]{2C2F33} \textbf{4.75}} &
  {\color[HTML]{2C2F33} \textbf{96.74}} \\
\multicolumn{1}{c|}{\cellcolor[HTML]{FFFFFF}} &
  $\Delta_{Base}$ &
  {\color[HTML]{FE0000} +0.29} &
  {\color[HTML]{FE0000} +4.76} &
  {\color[HTML]{FE0000} +0.32} &
  {\color[HTML]{FE0000} +5.30} &
  {\color[HTML]{FE0000} +0.20} &
  {\color[HTML]{FE0000} +2.95} &
  {\color[HTML]{FE0000} +0.26} &
  {\color[HTML]{FE0000} +2.14} \\ \bottomrule
\end{tabular}}
\caption{Main results on VQA out-domain generation.}
\label{tab:main-result-outdomain}
\end{table*}
\subsection{In-domain \& Out-domain Evaluations}
\label{sec:exp-main}

\textbf{In-domain VAQ Evaluation}. For the three in-domain test datasets, in our proposed more unbiased evaluation scheme, Iter-W2S-RLAIF achieves the best generalization performance. Compared to the baseline model LLaVA-Hound-SFT, the average scores on WebVid, VIDAL, and ActivityNet have increased by 0.28, 0.25, and 0.26, respectively, with ratios improved by 4.15\%, 3.45\%, and 3.42\%, showed in  \cref{tab:main-result-indomain}. Relative to other baselines, Iter-W2S-RLAIF demonstrates high performance on these in-domain datasets, indicating that the multi-AI feedback preference dataset construction pipeline  and the iterative DPO paradigm combined with parameters extrapolation  can improve the MLLMs average capability in in-domain VQA tasks.

It is noteworthy that for the two public models, LLaVA-Hound-SFT and LLaVA-Hound-DPO, in the original paper’s evaluation (the prompts used for evaluation refer to  \cref{fig:hound-dpo-eval}, LLaVA-Hound-DPO demonstrates score improvements over LLaVA-Hound-SFT by 4.91\%, 5.56\%, and 3.78\% on the three in-domain test datasets. However, under our proposed evaluation, the improvements are only 1.11\%, 1.16\%, and 0.68\%. This discrepancy arises because the original evaluation is lack of visual information input, and LLaVA-Hound-DPO merely aligns more closely with the groundtruth, leading to biased assessment of MLLMs VQA capability.

\begin{table*}[htp]
\centering
\resizebox{0.95\textwidth}{!}{
\begin{tabular}{
>{\columncolor[HTML]{FFFFFF}}c 
>{\columncolor[HTML]{FFFFFF}}c 
>{\columncolor[HTML]{FFFFFF}}c 
>{\columncolor[HTML]{FFFFFF}}c 
>{\columncolor[HTML]{FFFFFF}}c 
>{\columncolor[HTML]{FFFFFF}}c 
>{\columncolor[HTML]{FFFFFF}}c 
>{\columncolor[HTML]{FFFFFF}}c 
>{\columncolor[HTML]{FFFFFF}}c 
>{\columncolor[HTML]{FFFFFF}}c }
\toprule
\textbf{Base Model} &
  \textbf{Data} &
  \textbf{Iterative?} &
  \textbf{EXPO?} &
  \multicolumn{6}{c}{\cellcolor[HTML]{FFFFFF}\textbf{In-Domain}} \\ \hline
 &
   &
   &
  \multicolumn{1}{c|}{\cellcolor[HTML]{FFFFFF}} &
  \multicolumn{2}{c}{\cellcolor[HTML]{FFFFFF}\textbf{WebVid}} &
  \multicolumn{2}{c}{\cellcolor[HTML]{FFFFFF}\textbf{VIDAL}} &
  \multicolumn{2}{c}{\cellcolor[HTML]{FFFFFF}\textbf{ActivityNet}} \\
 &
   &
   &
  \multicolumn{1}{c|}{\cellcolor[HTML]{FFFFFF}} &
  Score &
  Ratio &
  Score &
  Ratio &
  Score &
  Ratio \\ \hline
H-SFT &
H-DPO-17k &
  \XSolidBrush &
  \multicolumn{1}{c|}{\cellcolor[HTML]{FFFFFF}\XSolidBrush} &
  {\color[HTML]{1F2329} 4.56} &
  {\color[HTML]{1F2329} 94.92} &
  4.36 &
  92.81 &
  4.44 &
  94.62 \\ \hline
\cellcolor[HTML]{FFFFFF} &
  \cellcolor[HTML]{FFFFFF} &
  \XSolidBrush &
  \multicolumn{1}{c|}{\cellcolor[HTML]{FFFFFF}\XSolidBrush} &
  {\color[HTML]{000000} 4.71} &
  95.80 &
  4.49 &
  93.08 &
  \multicolumn{1}{l}{\cellcolor[HTML]{FFFFFF}4.64} &
  95.98 \\
  
 {\cellcolor[HTML]{FFFFFF}H-SFT} &
  {\cellcolor[HTML]{FFFFFF}MMAIP-V} &
  \CheckmarkBold &
  \multicolumn{1}{c|}{\cellcolor[HTML]{FFFFFF}\XSolidBrush} &
  4.75 &
  96.64 &
  4.52 &
  93.60 &
  4.61 &
  94.92 \\
{} &
{} &
  \CheckmarkBold &
  \multicolumn{1}{c|}{\cellcolor[HTML]{FFFFFF}\CheckmarkBold} &
  {\color[HTML]{1F2329} \textbf{4.78}} &
  \multicolumn{1}{l}{\cellcolor[HTML]{FFFFFF}\textbf{97.67}} &
  {\color[HTML]{2C2F33} \textbf{4.52}} &
  \textbf{93.66} &
  {\color[HTML]{2C2F33} \textbf{4.64}} &
  {\color[HTML]{2C2F33} \textbf{96.13}} \\ \bottomrule
\end{tabular}}
\caption{Ablation study results on in-domain VQA generation. We conduct ablation experiments on DPO data, iterative DPO, and parameters extrapolation. Where ``\textbf{EXPO?}'' indicates whether parameters extrapolation is used, H-DPO-17k refers to the preference training datasets from the original LLaVA-Hound-DPO paper\cite{hound-dpo}, which is constructed based on the model's own responses and scoring from GPT-3.5 by inputting video subtitle proxy, with a total data volume of 17k. H-SFT represent the LLaVA-Hound-SFT~\cite{hound-dpo}.}
\label{tab:ablation-in-domain}
\end{table*}

\textbf{Out-domain VQA Evaluation.} For out-of-domain VQA generalization, Iter-W2S-RLAIF showes an average score improvement over the baseline model LLaVA-Hound-SFT by 0.29, 0.32, 0.20, and 0.26 on the four evaluation sets SSV2, MSRVTT, MSVD, and TGIF, respectively, with ratios improving by 4.76\%, 5.3\%, 2.95\%, and 3.26\%, showed in  \cref{tab:main-result-outdomain}. It also  outperform other baselines, indicating that Iter-W2S-RLAIF maintains strong out-of-domain generalization capability and exhibits  improvement in handling different types of VQA tasks.

Similar to the in-domain evaluation, the out-of-domain improvement margin under this evaluation framework is lower than that in the original paper. For instance, in the LLava-Hound-DPO original paper, the score on TGIF improved by 1.6\%, whereas in this evaluation, it decreased by 0.4\%. This further confirms that evaluating the relationship between MLLMs  response and groundtruth can introduce certain biases.

\begin{figure}[ht]
	\centering
	\begin{minipage}[c]{0.49\textwidth}
		\centering
		\includegraphics[width=1\textwidth]{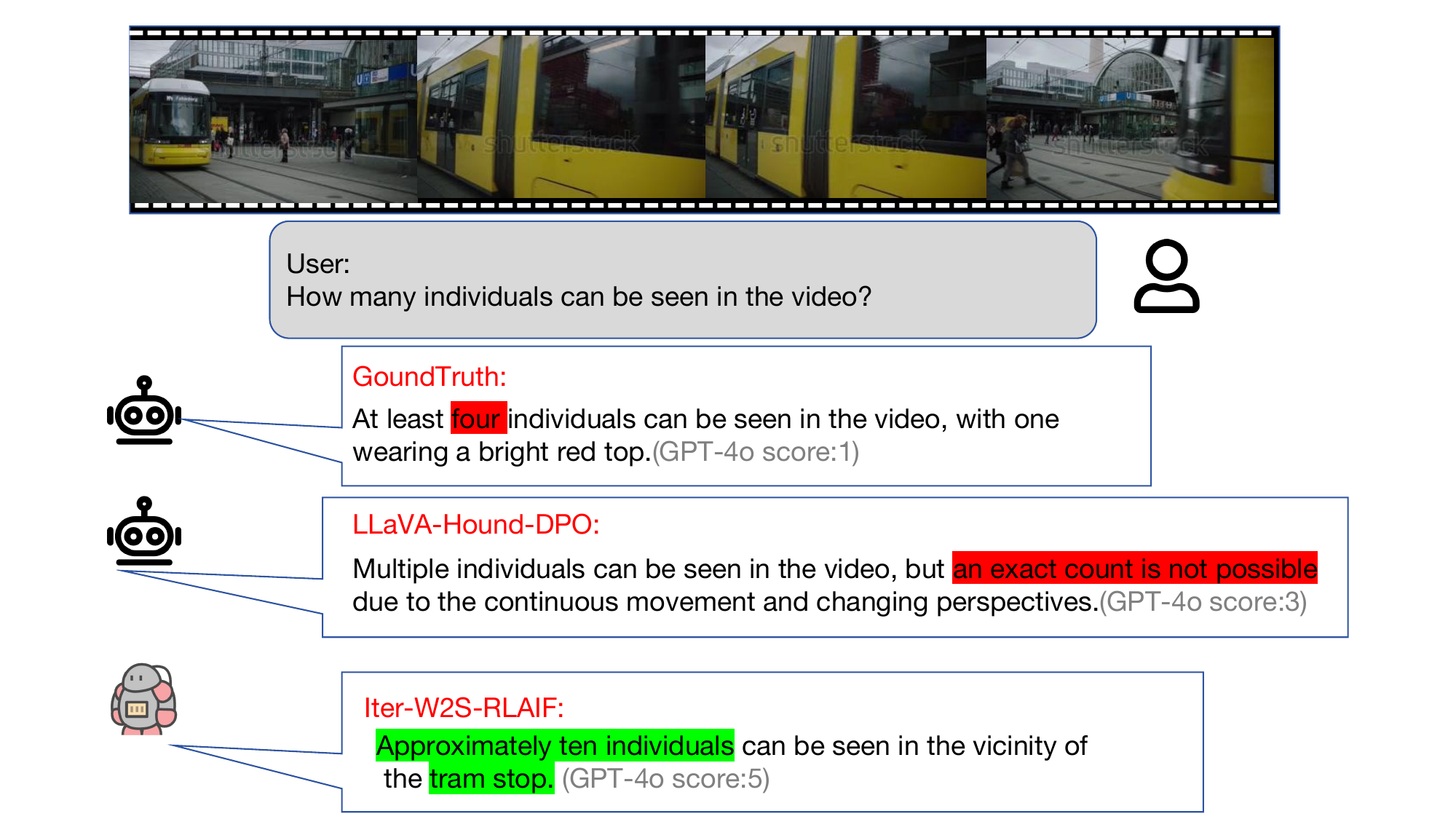}
		\subcaption{\textbf{WebVid}}
		\label{fig:case_study_webvide}
	\end{minipage} 
	\begin{minipage}[c]{0.49\textwidth}
		\centering
		\includegraphics[width=\textwidth]{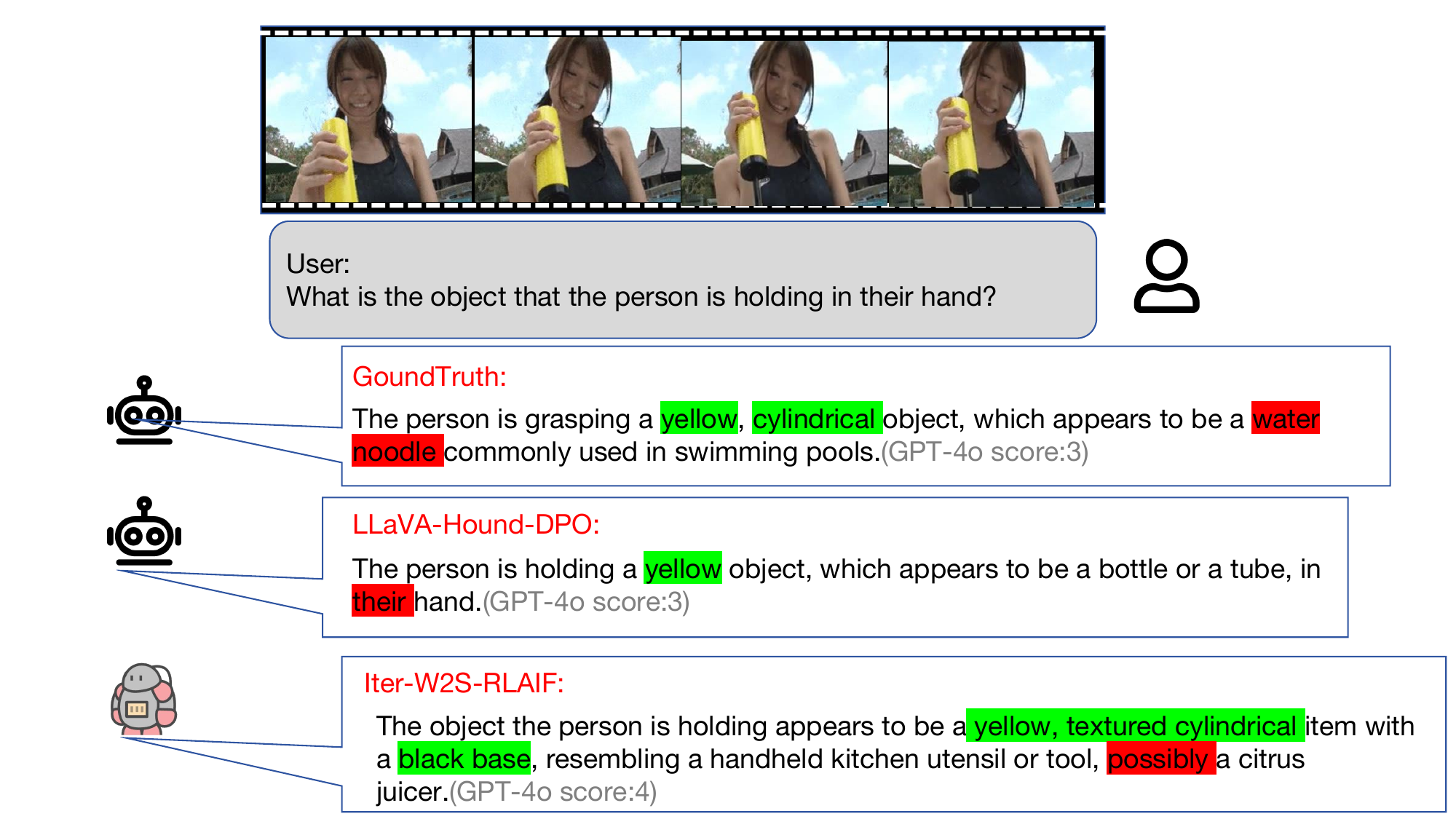}
		\subcaption{\textbf{TGIF}}
		\label{fig:case_study_tgif}
	\end{minipage} 
	\caption{Case Study on in-domain and out-domain VQA tasks.}

	\label{fig:case_study}
\end{figure}

\subsection{Ablation Study}\label{sec:abla}
\subsubsection{MMAIP-V}

\textbf{ The importance of MMAIP-V for VQA preference alignment.} Referring to   \cref{tab:ablation-in-domain}, \cref{tab:ablation-out-domain}, the results demonstrate that the video question-answering preference data constructed based on the MLLMs zoo is effective and is the main factor contributing to the improvement in VQA alignment performance. This is primarily due to the diversity of negative responses and the high quality of positive responses in MMAIP-V.

 \subsubsection{\textbf{The Fine-Grained Ablation Study of MMAIP-V.}}
 
\textbf{ Experiment Setup.} We modified the construction process of MMAIP-V and designed the following variants: 1) \textbf{2-MMAIP-V}: Using only Qwen2-VL-72B-instruct and Qwen2-VL-2B-instruct to build the preference dataset. 2) \textbf{2-MMAIP-V-P}: Based on the score distribution of Qwen2-VL-72B-instruct and Qwen2-VL-2B-instruct on MMAIP-V we can a prior consider Qwen2-VL-72B-instruct as having greater capability than Qwen2-VL-2B-instruct. Therefore, treating all responses from Qwen2-VL-72B-instruct as positive responses and all responses from Qwen2-VL-2B-instruct as negative responses. 3) \textbf{2-MMAIP-V-R}: Without referring to GPT-4's scores, randomly treating the responses from Qwen2-72B-instruct and Qwen2-VL-2B-instruct as positive and negative responses. 4)  \textbf{4-MMAIP-V-AW}: Using the lowest score answer as rejected answer and uniformly selecting other answer as chosen answer. 5)  \textbf{4-MMAIP-V-BA}: Using the highest score answer as chosen answer and uniformly selecting other answer as rejected answer. Due to the inconsistent amounts of preference dataset with these variant methods, we uniformly sample 8k entries from each dataset for more reasonable comparison. Neither iterative DPO or parameters extrapolation is performed.

\begin{table}[ht]
\centering
\resizebox{0.5\textwidth}{!}{
\begin{tabular}{ccccc}
\toprule
  \textbf{Dataset} &
  \multicolumn{2}{c}{\textbf{In-Domain}} &
  \multicolumn{2}{c}{\textbf{Out-Domain}} \\ \hline
    &
  \multicolumn{2}{c|}{\textbf{WebVid}} &
  \multicolumn{2}{c}{\textbf{SSV2}} \\
   &
  Score &
  \multicolumn{1}{c|}{Ratio} &
  Score &
  Ratio \\ \hline
  4-MMAIP-V\dag  &
  \textbf{4.62} &
  \multicolumn{1}{c|}{\textbf{94.51}} &
  \textbf{4.39} &
  \textbf{94.00} \\ \hline

   4-MMAIP-V-AW  &
  \begin{tabular}[c]{@{}c@{}}4.53\\ (\textcolor{red}{-0.09})\end{tabular} &
  \multicolumn{1}{c|}{\begin{tabular}[c]{@{}c@{}}93.23\\ (\textcolor{red}{-1.28})\end{tabular}} &
  \begin{tabular}[c]{@{}c@{}}4.23\\ (\textcolor{red}{-0.16})\end{tabular} &
  \begin{tabular}[c]{@{}c@{}}92.66\\ (\textcolor{red}{-1.34})\end{tabular} \\
  
  4-MMAIP-V-BA  &
  \begin{tabular}[c]{@{}c@{}}4.58\\ (\textcolor{red}{-0.04})\end{tabular} &
  \multicolumn{1}{c|}{\begin{tabular}[c]{@{}c@{}}94.01\\ (\textcolor{red}{-0.50})\end{tabular}} &
  \begin{tabular}[c]{@{}c@{}}4.32\\ (\textcolor{red}{-0.07})\end{tabular} &
  \begin{tabular}[c]{@{}c@{}}93.76\\ (\textcolor{red}{-0.24})\end{tabular} \\ \hline

  2-MMAIP-V  &
  \begin{tabular}[c]{@{}c@{}}4.55\\ (\textcolor{red}{-0.07})\end{tabular} &
  \multicolumn{1}{c|}{\begin{tabular}[c]{@{}c@{}}93.66\\ (\textcolor{red}{-0.85})\end{tabular}} &
  \begin{tabular}[c]{@{}c@{}}4.25\\ (\textcolor{red}{-0.14})\end{tabular} &
  \begin{tabular}[c]{@{}c@{}}93.21\\ (\textcolor{red}{-0.79})\end{tabular} \\ 
  2-MMAIP-V-P &
  \begin{tabular}[c]{@{}c@{}}4.52\\ (\textcolor{red}{-0.10})\end{tabular} &
  \multicolumn{1}{c|}{\begin{tabular}[c]{@{}c@{}}93.32\\ (\textcolor{red}{-1.19})\end{tabular}} &
  \begin{tabular}[c]{@{}c@{}}4.21\\ (\textcolor{red}{-0.18})\end{tabular} &
  \begin{tabular}[c]{@{}c@{}}92.56\\ (\textcolor{red}{-1.44})\end{tabular} \\ 
  2-MMAIP-V-R  &
  \begin{tabular}[c]{@{}c@{}}4.34\\ (\textcolor{red}{-0.28})\end{tabular} &
  \multicolumn{1}{c|}{\begin{tabular}[c]{@{}c@{}}90.12\\ (\textcolor{red}{-3.93})\end{tabular}} &
  \begin{tabular}[c]{@{}c@{}}4.02\\ (\textcolor{red}{-0.37})\end{tabular} &
  \begin{tabular}[c]{@{}c@{}}88.24\\ (\textcolor{red}{-5.76})\end{tabular} \\ \bottomrule
\end{tabular}}
\caption{The Fine-Grained Ablation Study of MMAIP-V. 4-MMAIP-V\dag denotes that we sample 8k entries from MMAIP-V.}
\label{tab:datasets-ablation}
\end{table}
\textbf{Analysis.} The results demonstrated in  \cref{tab:datasets-ablation} show that: 1) The diversity of responses comes from the responses distribution set sampling in MMAIP-V is crucial for VQA alignment. 2) If we utilize the MLLMs prior capability to construct pairs or randomly construct preference pairs instead of the scoring function evaluation in MMAIP-V, the preference learning may obtain noisy signals from datasets, which is harmful to MLLMs in VQA alignment. 3) Chosen answer or rejected answer comes from various distributions is harmful to alignment in MLLMs.

\subsubsection{\textbf{Iterative DPO \& Parameters Extrapolation.} }
 The iterative DPO strategy shows positive effects except on the ActivityNet dataset. The parameters extrapolation method further enhanced the alignment capability of the LMM. The most significant improvement was observed on MSRVTT, with the score increasing by 0.07 and the ratio by 2.08\%. This indicates that the iterative DPO and parameters extrapolation paradigms can further help the MLLMs utilize the AI alignment signals in MMAIP-V, compared to the original DPO framework. More ablation study refers to  \cref{sec:abl-appendix}.

\subsection{Case Study}
\label{sec:case_study}
We conduct a case study in both in-domain and out-domain VQA evaluation, as shown in \cref{fig:case_study}. We demonstrate video frames (4 frames), the question, and three answers (groundtruth, the responses from LLaVA-Hound-DPO and Iter-W2S-RLAIF), respectively. 

In the in-domain WebVid evaluation, the groundtruth has encountered a serious counting error about the individual. LLaVA-Hound-DPO only observes that there are many individuals but could not give a definite answer. Iter-W2S-RLAIF correctly counts the number of individuals and notices the individuals in the distance demonstrating the consistency and accuracy of model response and visual features in VQA tasks. This also further confirms the unreliability of groundtruth. In the out-domain TGIF evaluation, Iter-W2S-RLAIF gives four key characteristics of the object, which indicates that Iter-W2S-RLAIF is able to pay more attention to detailed vision features.

\section{Conclusion}

The lack of and low quality of VQA preference datasets severely harm the preference learning of MLLMs. This work proposes  a high-quality VQA preference dataset, called MMAIP-V, which is beneficial to preference learning. Additionally, we propose Iter-W2S-RLAIF, a framework that gradually enhances MLLMs' alignment capabilities through iteratively updating  the reference model and performing parameter extrapolation. Further, we  propose a new scheme, which integrates visual information and employs more fine-grained evaluations. We believe that the proposed automatic VQA preference data generation pipeline based on multi-AI feedback and weak-to-strong offline reinforcement learning framework can greatly promote future work in the MLLMs alignment.
\label{sec:conclusion}

{
    \bibliographystyle{unsrt}
    \bibliography{main}
}

\clearpage
\setcounter{page}{1}
\appendix

\begin{figure}[ht]
    \centering
    \includegraphics[width=0.75\linewidth]{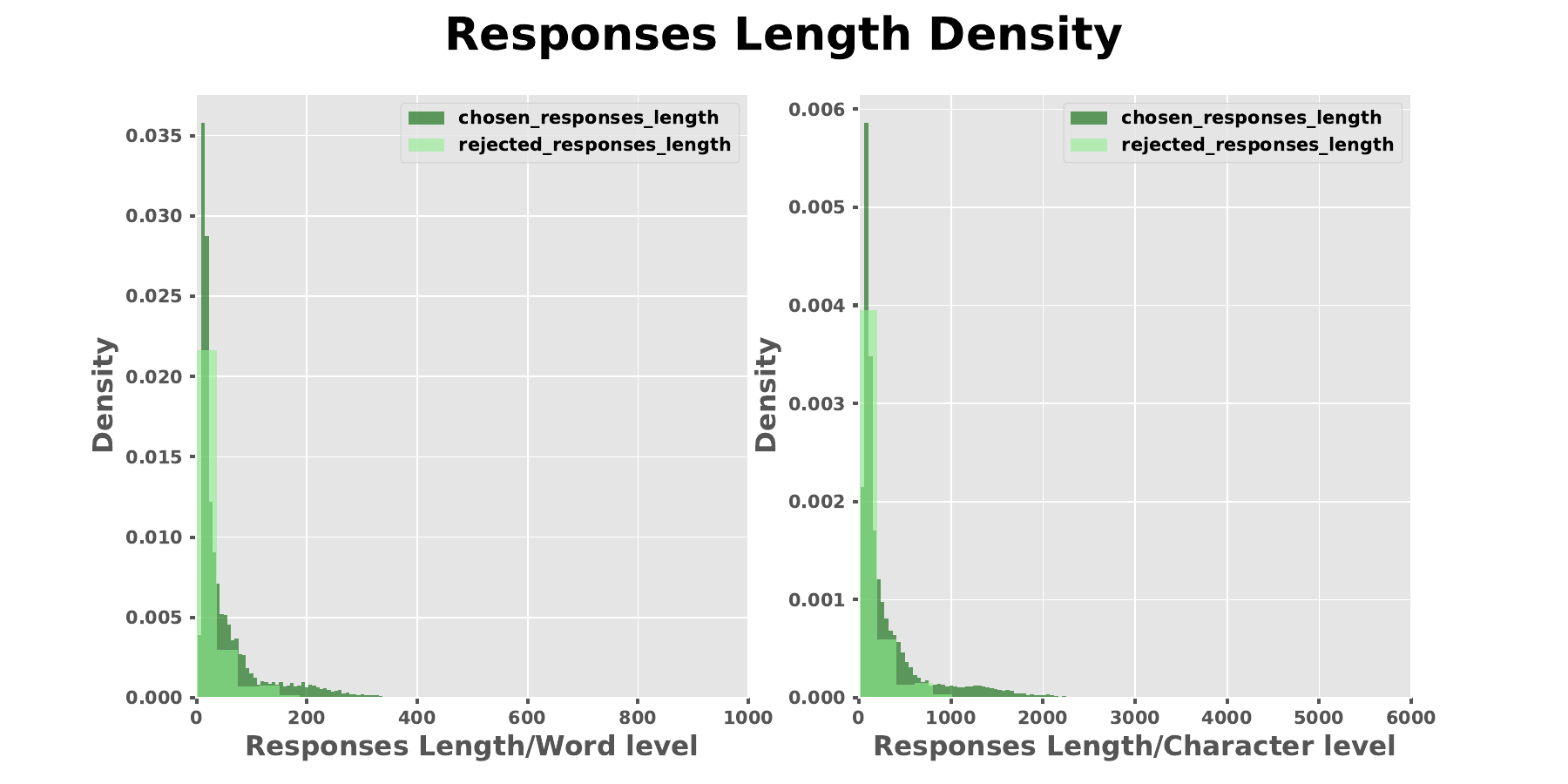}
    \caption{MMAIP-V responses length distribution.}
    \label{fig:length distribution.}
\end{figure}
\begin{figure}[ht]
    \centering
    \includegraphics[width=0.5\linewidth]{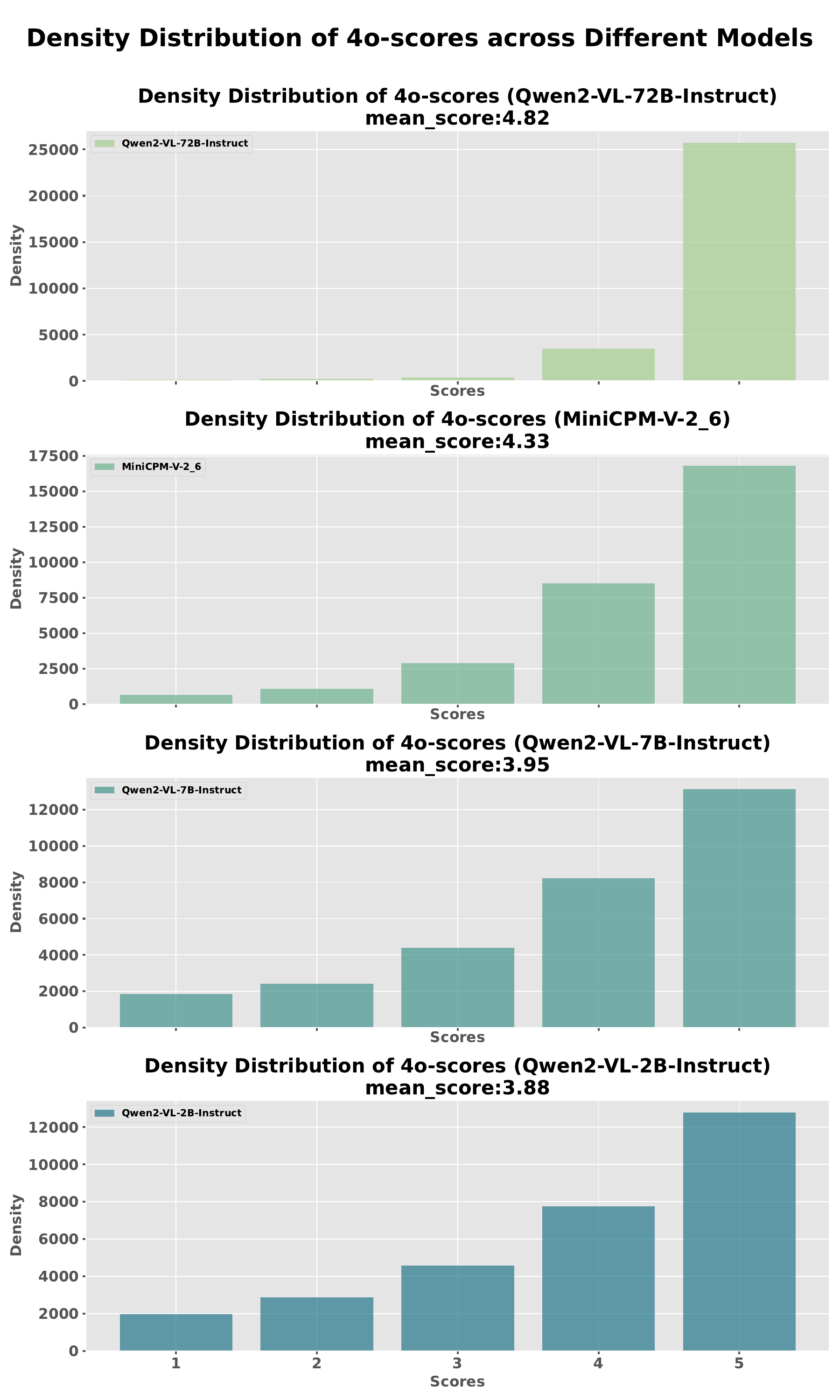}
    \caption{GPT-4o score distributions of responses from various models in the MLLMs zoo.}
    \label{fig:score-distribution}
\end{figure}

\begin{figure}[ht]
    \centering
    \includegraphics[width=1\linewidth]{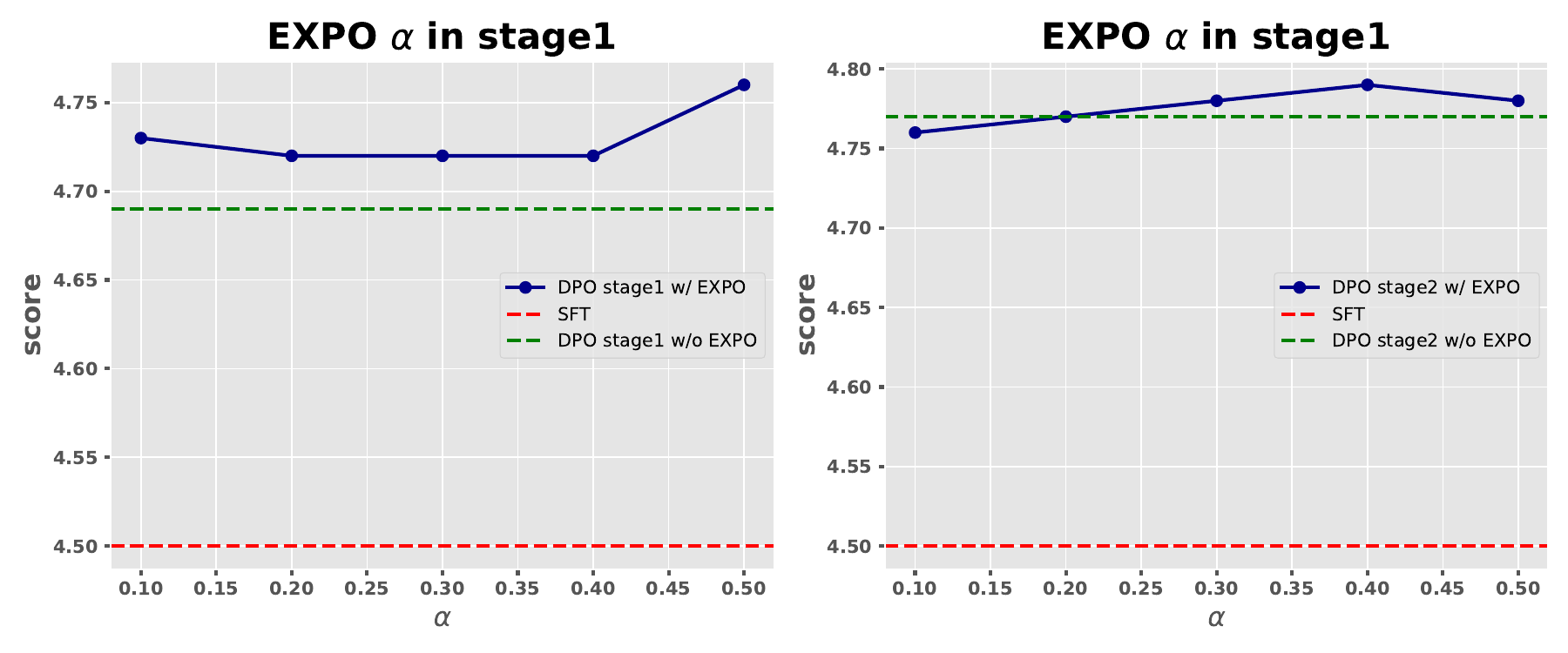}
    \caption{$\alpha$ in Parameters Extrapolation.}
    \label{fig:abl-alpha}
\end{figure}

\begin{table*}[hpt]
\centering
\begin{tabular}{ccccc}
\toprule
Model                 & Size                      & Vision Encoder                                    & { LLM} & Projector                        \\ \hline
LLaVA-Hound-SFT \cite{hound-dpo}     & { 7B} & { LanguageBind-video \cite{languagebind}}         & vicuna-7b                  & { MLP2-GeLU} \\
LLaVA-Hound-DPO  \cite{hound-dpo}     & { 7B} & { LanguageBind-video}         & vicuna-7b                  & { MLP2-GeLU} \\
VLM-RLAIF  \cite{vlm-rlaif}           & { 7B} & clip-vit-large-patch14-336 \cite{clip}                       & vicuna-7b                  & { MLP2-GeLU} \\
LLaMA-VID  \cite{llama-vid}        & { 7B} & EVA-G  \cite{eva}                                           & vicuna-7b                  & { QFormer}\cite{qformer}   \\
Video-LLaVA  \cite{video-llava}         & { 7B} & { LanguageBind-video}         & vicuna-7b                  & { MLP2-GeLU} \\
LLaVA-Next-Video-base  \cite{llavanextvideo}& { 7B} & { clip-vit-large-patch14-336} & vicuna-7b                  & { MLP2-GeLU} \\
LLaVA-Next-Video-DPO \cite{llavanextvideo} & { 7B} & { clip-vit-large-patch14-336} & vicuna-7b                  & { MLP2-GeLU} \\
VideoChatGPT \cite{video-chatgpt}         & { 7B} & { clip-vit-large-patch14}     & vicuna-7b                  & { Linear}    \\
VideoLLaMA2-base  \cite{videollama2}      & { 7B} & { clip-vit-large-patch14-336} & Mistral-7B-Instruct-v0.2   & { MLP2-GeLU} \\
VideoLLaMA2-chat  \cite{videollama2}    & { 7B} & { clip-vit-large-patch14-336} & Mistral-7B-Instruct-v0.2   & { MLP2-GeLU} \\ \bottomrule
\end{tabular}
\caption{Baselines model detailed information.}
\label{tab:baseline}
\end{table*}

\begin{algorithm}
\caption{MMAIP-V Construction \& Iter-W2S-RLAIF} 
\label{alg1}
\begin{algorithmic}
\REQUIRE MMLMs Zoo $\{\mathcal{M}_{i}\}_{i=1}^K$, Scoring Function $R$, Policy Model $\pi_{\theta_0}$,Video distribution $\mathcal{V}$, Question distribution $\mathcal{X}$, Video-Question Datasets $\{(v_i,x_i) | v_i\sim \mathcal{V}, x_i\sim \mathcal{X}\}_{i=1}^N$, iteration times $T$, learning rate $\gamma$, parameters extrapolation hyperparamter $\alpha$.
\ENSURE Aligned Policy Model $\pi_{\theta_T}$. \\

\textbf{\textcolor{red}{\#\# MMAIP-V}}
\STATE $\mathcal{D}_{\text{DPO}}^{VT}=\phi$;
\FOR{$i$ from $1$ to $N$}
  \FOR{$j$ from $i$ to $K$}
    \STATE $y_j\sim \mathcal{M}_j(y|(v_i,x_i))$; \textcolor{blue}{\COMMENT{Sampling Response.}}
    \STATE $r_j = R(y_j;(v_i,x_i))$; \textcolor{blue}{\COMMENT{Automatic Scoring.}}
  \ENDFOR
  \STATE $y_w = \arg\max_{y_j} r_j$;
  \STATE $y_l = \arg\min_{y_j} r_j$;
  \STATE$ (v_i,q_i,y_w,y_l)$ append to $\mathcal{D}_{\text{DPO}}^{VT}$.
\ENDFOR

\textbf{\textcolor{red}{\#\# Iter-W2S-RLAIF}}
\STATE Segment $\mathcal{D}_{\text{DPO}}^{VT}$ into $\{\mathcal{D}_i\}_{i=1}^T$ evenly;
\STATE Initialize $\pi_{ref}=\pi_{\theta_0}$.
\FOR{$t$ from 1 to $T$}
    \STATE Calculate $\mathcal{L}_{DPO}^t$ referring to  \cref{equ:dpo-loss} on $\mathcal{D}_t$;
    \STATE $\theta^\prime_{t} = \theta_{t-1} - \gamma\nabla_{\theta_{t-1}}\mathcal{L}_{DPO}^T$ ;
    \STATE $\theta_t = \theta^\prime_{t} + \alpha(\theta^\prime_{t}-\theta_{0})$;\textcolor{blue}{\COMMENT{Parameters Extrapolation.}}
    \STATE $\pi_{ref} = \pi_{\theta_t}$. \textcolor{blue}{\COMMENT{Iterative DPO.}}
\ENDFOR

\RETURN $\pi_{\theta_T}$
\end{algorithmic}
\end{algorithm}

\begin{table*}[ht]
\centering
\begin{tabular}{
>{\columncolor[HTML]{FFFFFF}}c 
>{\columncolor[HTML]{FFFFFF}}c 
>{\columncolor[HTML]{FFFFFF}}c 
>{\columncolor[HTML]{FFFFFF}}c 
>{\columncolor[HTML]{FFFFFF}}c 
>{\columncolor[HTML]{FFFFFF}}c 
>{\columncolor[HTML]{FFFFFF}}c 
>{\columncolor[HTML]{FFFFFF}}c 
>{\columncolor[HTML]{FFFFFF}}c 
>{\columncolor[HTML]{FFFFFF}}c 
>{\columncolor[HTML]{FFFFFF}}c 
>{\columncolor[HTML]{FFFFFF}}c }
\toprule
\textbf{Base Model} &
  \textbf{Data} &
  \textbf{Iterative?} &
  \textbf{EXPO?} &
  \multicolumn{8}{c}{\cellcolor[HTML]{FFFFFF}\textbf{Out-Domain}} \\ \hline
 &
   &
   &
  \multicolumn{1}{c|}{\cellcolor[HTML]{FFFFFF}} &
  \multicolumn{2}{c}{\cellcolor[HTML]{FFFFFF}\textbf{SSV2}} &
  \multicolumn{2}{c}{\cellcolor[HTML]{FFFFFF}\textbf{MSRVTT}} &
  \multicolumn{2}{c}{\cellcolor[HTML]{FFFFFF}\textbf{MSVD}} &
  \multicolumn{2}{c}{\cellcolor[HTML]{FFFFFF}\textbf{TGIF}} \\
 &
   &
   &
  \multicolumn{1}{c|}{\cellcolor[HTML]{FFFFFF}} &
  Score &
  Ratio &
  Score &
  Ratio &
  Score &
  Ratio &
  Score &
  Ratio \\ \hline
\cellcolor[HTML]{FFFFFF} &
\textbf{H-DPO-17k} &
  \XSolidBrush &
  \multicolumn{1}{c|}{\cellcolor[HTML]{FFFFFF}\XSolidBrush} &
  {\color[HTML]{1F2329} 4.28} &
  {\color[HTML]{1F2329} 94.24} &
  4.24 &
  89.86 &
  4.44 &
  93.35 &
  4.53 &
  95.60 \\
\cellcolor[HTML]{FFFFFF} &
  \cellcolor[HTML]{FFFFFF} &
  \XSolidBrush &
  \multicolumn{1}{c|}{\cellcolor[HTML]{FFFFFF}\XSolidBrush} &
  {\color[HTML]{000000} 4.44} &
  94.33 &
  4.45 &
  91.49 &
  4.58 &
  93.80 &
  4.70 &
  96.04 \\
  
{\cellcolor[HTML]{FFFFFF}H-SFT} &
{\cellcolor[HTML]{FFFFFF}MMAIP-V} &
  \CheckmarkBold &
  \multicolumn{1}{c|}{\cellcolor[HTML]{FFFFFF}\XSolidBrush} &
  4.47 &
  95.30 &
  4.45 &
  91.63 &
  4.60 &
  94.39 &
  4.71 &
  96.08 \\
    & &
  \CheckmarkBold &
  \multicolumn{1}{c|}{\cellcolor[HTML]{FFFFFF}\CheckmarkBold} &
  \multicolumn{1}{l}{\cellcolor[HTML]{FFFFFF}{\color[HTML]{1F2329} \textbf{4.51}}} &
  \multicolumn{1}{l}{\cellcolor[HTML]{FFFFFF}{\color[HTML]{1F2329} \textbf{95.48}}} &
  {\color[HTML]{2C2F33} \textbf{4.52}} &
  \multicolumn{1}{l}{\cellcolor[HTML]{FFFFFF}\textbf{93.22}} &
  \multicolumn{1}{l}{\cellcolor[HTML]{FFFFFF}{\color[HTML]{2C2F33} \textbf{4.61}}} &
  {\color[HTML]{1F2329} \textbf{94.66}} &
  {\color[HTML]{2C2F33} \textbf{4.75}} &
  {\color[HTML]{2C2F33} \textbf{96.74}} \\ \bottomrule
\end{tabular}
\caption{Ablation study results on out-domain VQA generation.}
\label{tab:ablation-out-domain}
\end{table*}

\begin{table*}[ht]
\begin{tabular}{cccccc}\toprule
 &
  $\mathbb{E}|y_w|$ &
  $\mathbb{E}|y_l|$ &
  \# of frames &
  Source $y_w$ &
  Source $y_l$ \\ \hline
MMAIP-V &
  298.28 &
  253.48 &
  1386908 &
  \begin{tabular}[c]{@{}c@{}}MiniCPM-V-2.6: 5428\\ Qwen2-VL-2B-Instruct: 2379\\ Qwen2-VL-7B-Instruct:5307\\ Qwen2-VL-72B-Instruct:10914\end{tabular} &
  \begin{tabular}[c]{@{}c@{}}MiniCPM-V-2.6: 7217\\ Qwen2-VL-2B-Instruct:7062\\ Qwen2-VL-7B-Instruct:8895\\ Qwen2-VL-72B-Instruct:854 

  \end{tabular}\\ \bottomrule
\end{tabular}
\caption{MMAIP-V statistics details.}
\label{tab:MMAIP-stat}
\end{table*}

\section{MMAIP-V Details}
\label{sec:MMAIP-V-datails}
\subsection{MMAIP-V Statistic Summary}
We summarize the features about MMAIP-V, including the mean of chosen answers $\mathbb{E}|y_w|$, the mean of rejected answer $\mathbb{E}|y_l|$, the total number of video frames and the source of chosen answers and rejected answer, seeing \cref{tab:MMAIP-stat}.

\subsection{Length distribution of positive and negative responses}
We analyze the length distribution of positive and negative responses at both the character level and the word level on MMAIP-V. It demonstrates that at both the character level and the word level, positive responses exhibit a peak and long-tailed distribution. This indicates that compared to negative responses, positive responses require longer descriptions for some prompts and videos inputs. Such responses are often more detailed, comprehensive, and thorough, but also increase the risk of the model's answers containing hallucinations. However, this kind of risk can be effectively mitigated using a scoring function, like GPT-4o we use in this work.

\section{Experiment Details}\label{sec:exp-detail}
\subsection{Model details}\label{sec:model-detail}
The base model used in this experiment is \texttt{Llava-Hound-SFT} \cite{hound-dpo}\footnote{Publicly available at \href{https://huggingface.co/ShareGPTVideo/LLaVA-Hound-SFT}{LLaVA-Hound-SFT}}.
This model leverages the MLLMs model proposed by \texttt{Video-LLaVA}~\cite{video-llava} and is fine-tuned using 300k data constructed from video proxy subtitles. The language model used is \texttt{vicuna-7b-v1.5}~\cite{vicuna}, the vision model employs the encoder from \texttt{LanguageBind}\cite{languagebind} for both video and images, and the multimodal projector is a two-layer \texttt{MLP}. The total number of model parameters is 7B.

\subsection{Baselines Details}\label{sec:baselines-detail}
For a fairer comparison, we select the latest MLLMs of the same size and language understanding capability, including models that have undergone Pretrained, SFT, or RLHF/RLAIF training. These include:  dModels at the \textit{Pretained} stage: LLaMA-VID\cite{llama-vid}, Video-LLaVA\cite{video-llava}, LLava-Next-Video-base\cite{llavanextvideo}, VideoChatGPT\\~\cite{video-chatgpt}, VideoLLaMA2-base\cite{videollama2}.
Models at the \textit{SFT} stage: LLaVA-Hound-SFT\cite{hound-dpo}, VideoLLaMA2-chat\cite{hound-dpo}.
Models at the \textit{RLHF/RLAIF} stage: LLaVA-Hound-DPO\cite{hound-dpo}, VLM-RLAIF\cite{vlm-rlaif}, LLava-Next-Video-DPO\cite{llavanextvideo}.

Detailed information on the baselines is presented in  \cref{tab:baseline}, including the number of model parameters, the type of visual encoder, the type of LLM and the type of projector.

\subsection{Training Details}
For all videos, we uniformly extract 8 frames as input for the model's visual encoder. For the iterative DPO training, we set 2 rounds for iterations and evenly split the constructed 24k preference data into two parts, denoted as $\mathcal{D}_1$ and $\mathcal{D}_2$. For each round of DPO training, we set the $\beta$ in the loss function to $0.1$, selected Adamw ~\cite{adamw} as the optimizer, and set the learning rate to $5e-7$. For parameters extrapolation, we perform parameters extrapolation at the end of both the first and second stages. Regarding the coefficient $\alpha$ for parameters extrapolation, we searched within $\{0.1, 0.2, 0.3, 0.4, 0.5\}$ and chose the optimal setting based on the \textbf{WebVid} validation datasets. All experiments are conducted on 8 Nvidia H800 80GB GPUs.

\section{Prompts}
\label{sec:prompts}

\section{Ablation Study Supplementary Material}
\label{sec:abl-appendix}
\subsection{ $\alpha$ in Parameters Extrapolation}
\label{sec:alpha-pe}
The detailed analysis is demonstrated in  \cref{sec:alpha-pe}. We experimented with different parameter extrapolation coefficients $\alpha$ at various stages of iterative DPO, including \{0.1, 0.2, 0.3, 0.4, 0.5\}. The left figure represents the results of parameter extrapolation in the first stage, while the right figure represents the results of parameter extrapolation in the second stage. The \textcolor{red}{red} line represents the model indicated by $\theta_0$ in parameter extrapolation, the \textcolor{green}{green} line represents the model indicated by $\theta_1$ in parameter extrapolation, and the \textcolor{blue}{blue} lines represent the interpolated models. It can be observed that in the first stage, the interpolated models consistently outperform the($\theta_1$ model, with the best performance achieved at $0.5$. In the second stage, the improvement in model alignment performance due to interpolation is not as significant as in the first stage, but with an optimal choice of $\alpha$, the interpolated models can still surpass the $\theta_1$ model.

\begin{figure*}[th]
    \centering
    \includegraphics[width=0.9\linewidth]{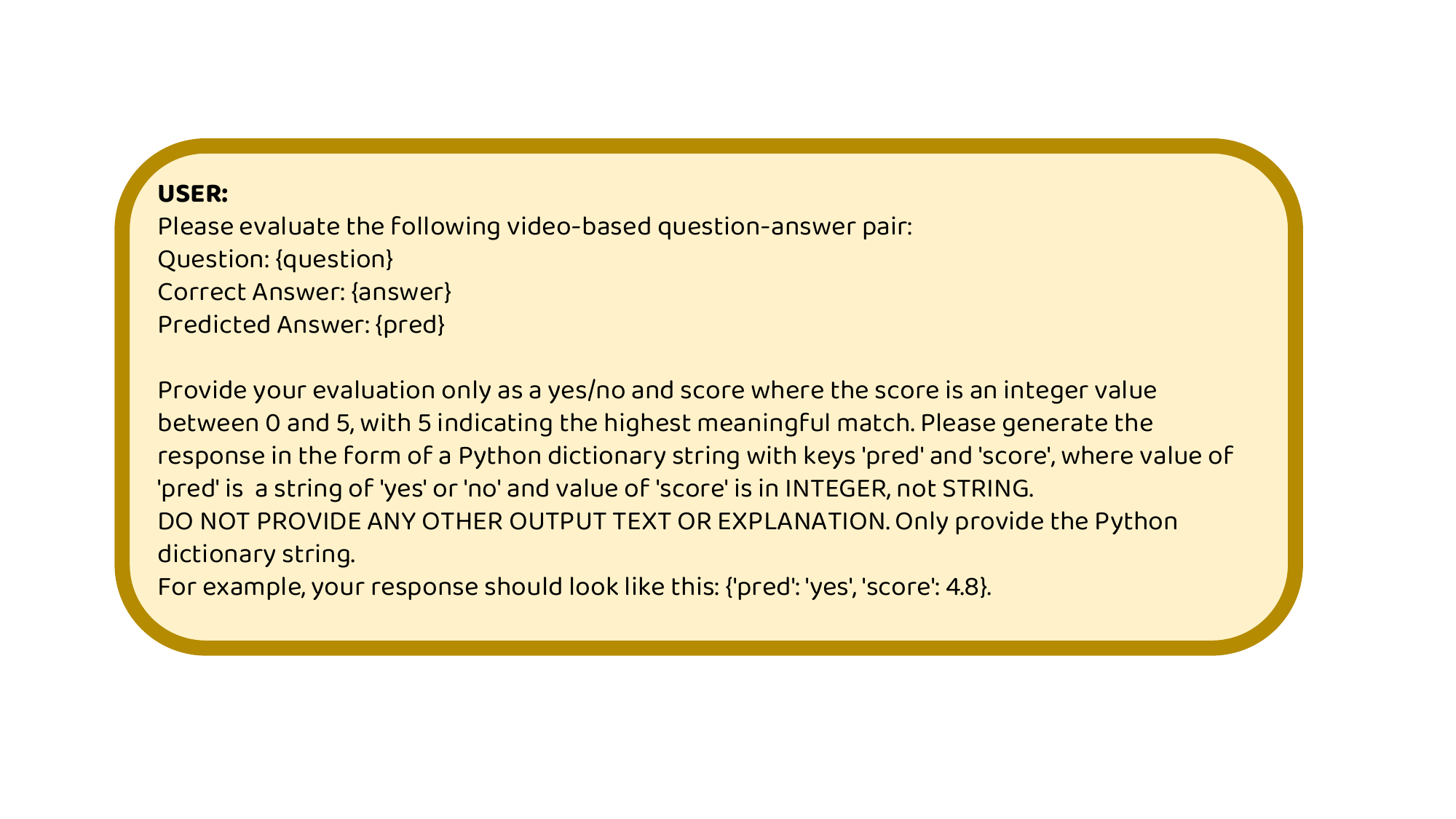}
    \caption{Previous VQA evaluation prompts.}
    \label{fig:old-eval-prompt}
\end{figure*}
\begin{figure*}[tph]
    \centering
    \includegraphics[width=0.9\linewidth]{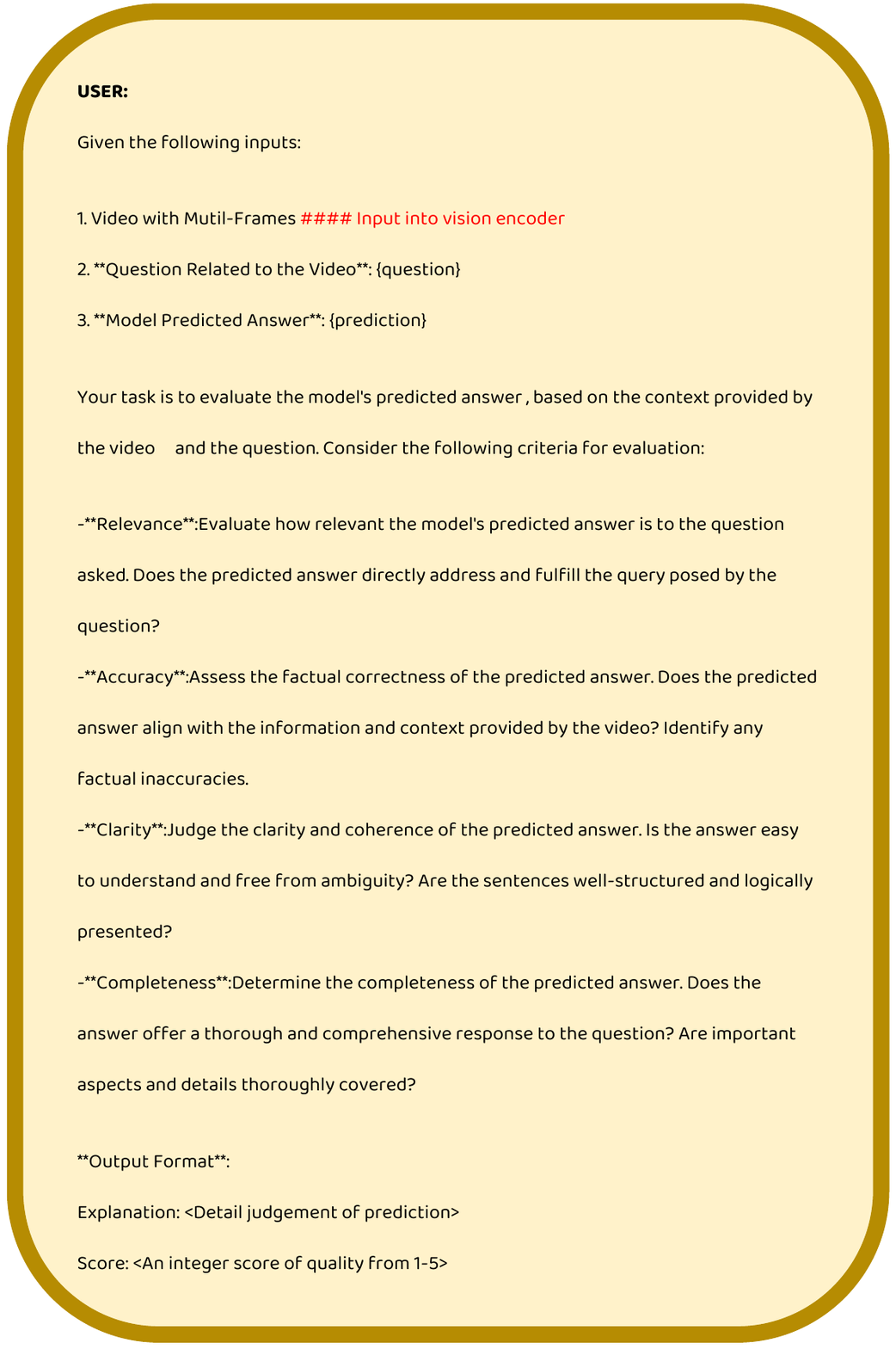}
    \caption{Proposed VQA evaluation prompts.}
    \label{fig:new-eval-prompt}
\end{figure*}

\begin{figure*}[th]
    \centering
    \includegraphics[width=0.9\linewidth]{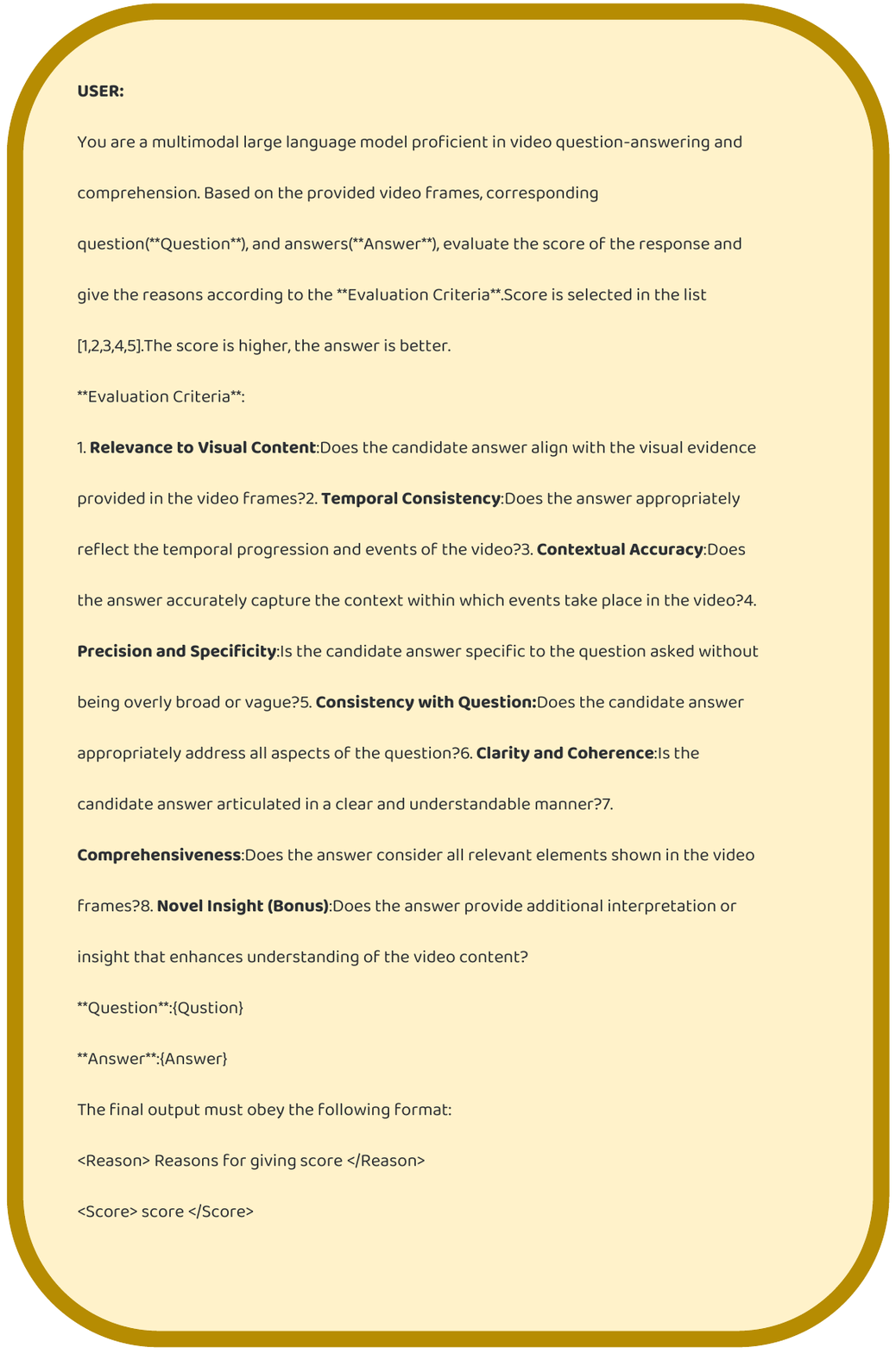}
    \caption{MLLMs zoo Responses scoring prompts.}
    \label{fig:infer_reward}
\end{figure*}

\begin{figure*}[th]
    \centering
    \includegraphics[width=0.9\linewidth]{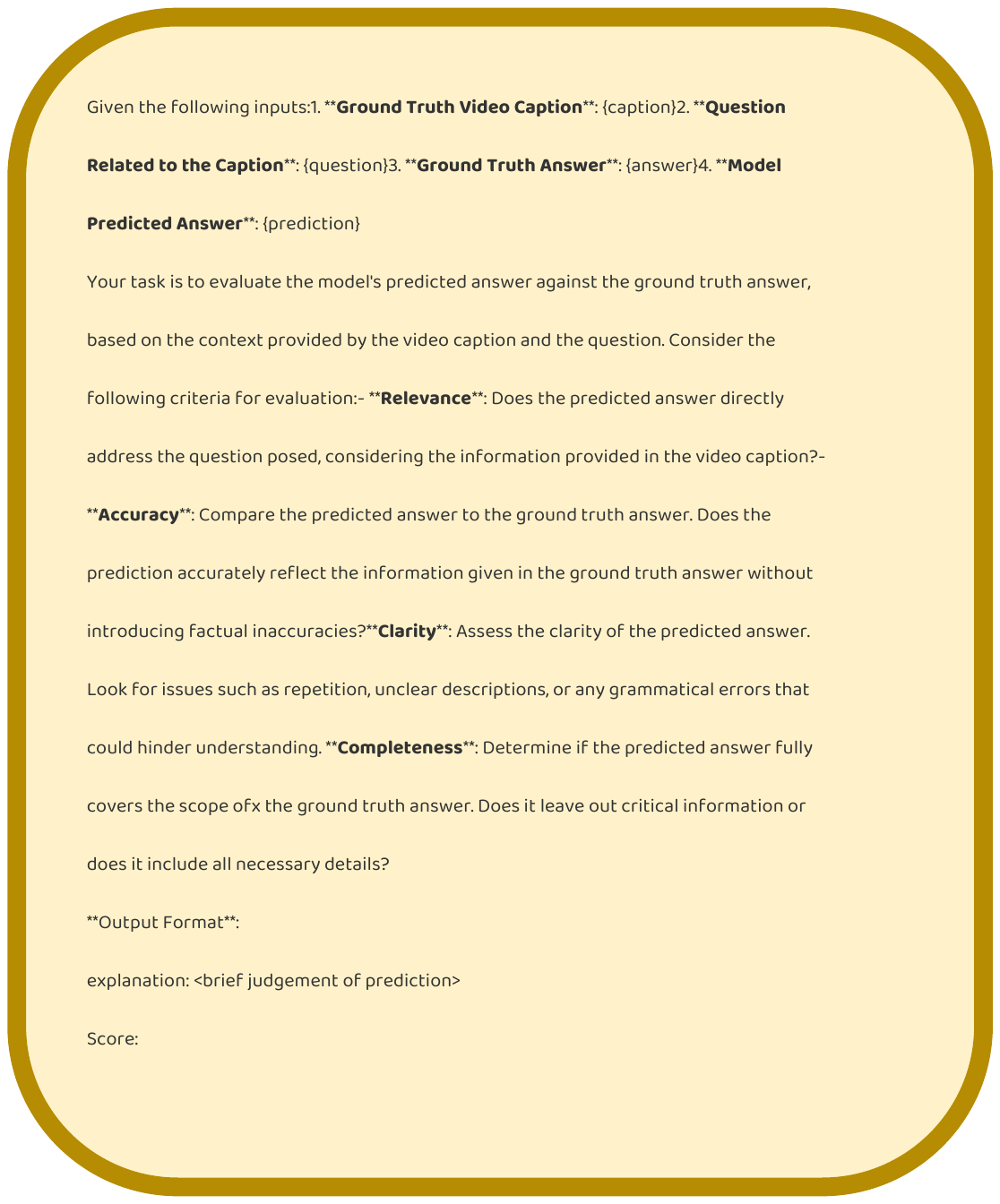}
    \caption{The evaluation prompts used in LLaVA-Hound-DPO\cite{hound-dpo}}
    \label{fig:hound-dpo-eval}
\end{figure*}

\end{document}